\documentclass{article}

\usepackage{arxiv}
\usepackage{upquote}

\usepackage[utf8]{inputenc} 
\usepackage[T1]{fontenc}    
\usepackage{hyperref}       
\usepackage{url}            
\usepackage{booktabs}       
\usepackage{amsfonts}       
\usepackage{amssymb}
\usepackage{nicefrac}       
\usepackage{microtype}      
\usepackage{lipsum}
\usepackage{graphicx}
\usepackage{mdframed}
\usepackage{amsmath}
\usepackage{float}
\usepackage{xcolor}

\usepackage{framed,color,verbatim}
\definecolor{shadecolor}{rgb}{.9, .9, .9}
\definecolor{cyan}{rgb}{.0, .99, .99}

\graphicspath{ {./images/} }

\usepackage{url}
\usepackage[pdftex]{epsfig}

\usepackage{color}
\newcommand{\remove}[1]{ }

\newenvironment{mycodebox}%
   {\snugshade\verbatim}%
   {\endverbatim\endsnugshade}

\newcommand{\mywarning}[1]{\colorbox{cyan}{\parbox{\textwidth}{{\bf A word of caution:} #1}}}

\title{CIRA Guide to Custom Loss Functions for Neural Networks in Environmental Sciences - Version 1}

\author{
Imme Ebert-Uphoff
\thanks{{\bf CIRA:} Cooperative Institute for Research in the Atmosphere, Colorado State University, Fort Collins, CO. 
\hspace{0.2cm} 
{\bf ECE:}  Electrical and Computer Engineering, Colorado State University, Fort Collins, CO.
\hspace{0.2cm} 
{\bf NOAA-GSL:} National Oceanic and Atmospheric Administration (NOAA), Global Systems Laboratory (GSL), Boulder, Colorado.  
\hspace{0.2cm} 
{\bf CS:} Computer Science, Colorado State University, Fort Collins, CO.
\hspace{0.2cm} 
{\bf CIRES:} Cooperative Institute for Research in Environmental Sciences, University of Colorado Boulder, Boulder, CO.}
\\ 
  CIRA, ECE\\
  \texttt{iebert@colostate.edu} \\
   \And
Ryan Lagerquist\\ 
  CIRA, NOAA-GSL\\
  \texttt{ralager@colostate.edu}
     \And
Kyle Hilburn\\ 
  CIRA\\
  \texttt{Kyle.Hilburn@colostate.edu}
     \And
Yoonjin Lee\\ 
  CIRA\\
  \texttt{Yoonjin.Lee@colostate.edu}
     \And
Katherine Haynes\\ 
  CIRA\\
  \texttt{Katherine.Haynes@colostate.edu}
     \And
Jason Stock\\
  CS\\
  \texttt{Jason.Stock@rams.colostate.edu}
     \And
Christina Kumler\\ 
  CIRES, NOAA-GSL\\
  \texttt{christina.e.kumler@noaa.gov}
     \And
Jebb Q.\ Stewart\\ 
  NOAA-GSL\\
  \texttt{jebb.q.stewart@noaa.gov}
}

\begin{document}
\maketitle
\begin{abstract}
Neural networks are increasingly used in environmental science applications.  Furthermore, neural network models are trained by minimizing a loss function, and it is crucial to choose the loss function very carefully for environmental science applications, as it determines what exactly is being optimized.  Standard loss functions do not cover all the needs of the environmental sciences, which makes it important for scientists to be able to develop their own custom loss functions so that they can implement many of the classic performance measures already developed in environmental science, including measures developed for spatial model verification.  However, there are very few resources available that cover the basics of custom loss function development comprehensively, and to the best of our knowledge none that focus on the needs of environmental scientists.
This document seeks to fill this gap by providing a guide on how to write custom loss functions targeted toward environmental science applications. Topics include the basics of writing custom loss functions, common pitfalls, functions to use in loss functions, examples such as fractions skill score as loss function, how to incorporate physical constraints, discrete and soft discretization, and concepts such as focal, robust, and adaptive loss.  
While examples are currently provided in this guide for Python with  
Keras and the TensorFlow backend, the basic concepts also apply to other environments, such as Python with PyTorch.  
Similarly, while the sample loss functions provided here are from meteorology, these are just examples of how to create custom loss functions.  Other fields in the environmental sciences have very similar needs for custom loss functions, e.g., for evaluating spatial forecasts effectively, and the concepts discussed here can be applied there as well.  All code samples are provided in a GitHub repository. 
\end{abstract}



\section{Why would environmental scientists care about custom loss functions?}

The use of neural networks in environmental science applications is growing at a rapid pace. In order to train a neural network one has to choose a cost function, called a {\it loss function} in the context of neural networks, which represents the error of the neural network. The neural network is then trained, i.e. its parameters are chosen through an iterative process, such that the loss function, and thus the error, is minimized. 
It is crucial to choose the loss function very carefully for environmental applications, as it determines what exactly the neural network is optimizing.  Many pre-defined loss functions exist. The most popular examples for regression (predicting a continuous value such as wind speed) include mean absolute error (MAE), mean squared error (MSE), and root mean squared error (RMSE). The most popular example for classification is cross-entropy. 
There are many other predefined loss functions, and their number keeps growing, but they do not cover everything environmental scientists care about, since they were developed for other applications.  In fact, environmental scientists have a long tradition of developing meaningful performance measures for forecasting tasks, such as for single-category forecasts (accuracy, frequency bias, probability of detection, success ratio, etc.);
for multi-category forecasts (Heidke score, Gerrity score, etc.);
for continuous forecasts (correlation, reliability diagram, etc.);
for probabilistic forecasts (reliability diagram, Brier score, etc.);
for spatial forecasts (neighborhood methods, such as fractions skill score, and scale decomposition, such as wavelet decomposition), and many others.  For an extremely comprehensive overview of these and other performance measures, see the guide of the
{\it WWRP/WGNE Joint Working Group on Forecast Verification Research} at \url{https://www.cawcr.gov.au/projects/verification/}.

However, it is not obvious which ones of those performance measures can easily be used in a neural network and how to do it, for the following reasons:
\begin{itemize}
\item 
    There are various limitations of what can be implemented in a neural network loss function. Functions must be differentiable and execute extremely quickly, which makes it tricky to implement custom loss functions.
\item
    The loss functions required by environmental scientists are unlike any loss functions typically used in computer science, and the community has not yet developed comprehensive resources, such as a large collection of customized loss functions.
\end{itemize}
The above reasons make the topic of loss functions a significant hurdle for practitioners striving to implement meaningful loss functions for their applications. 
We seek to close this gap here by providing comprehensive instructions, including many examples and discussion of common pitfalls, on how to code custom loss functions.
While the sample loss functions provided here are from meteorology, these are just examples of how to create custom loss functions.  Other fields in the environmental sciences have very similar needs for custom loss functions, e.g., for evaluating spatial forecasts effectively, and the concepts discussed here can be applied there as well.
Making it possible, and even easy, to use a variety of meaningful loss functions will go a long way to more effectively tune neural networks to focus on the types of performance criteria that are truly important in environmental science applications, thus helping the science community to make the most of neural networks for their applications.   
Scientists in other areas have invested in similar efforts, e.g., researchers in the medical imaging community have developed a collection of loss functions specifically for medical image segmentation \cite{ma2021loss}.  Loss function development for different purposes also remains a very active topic in computer science.  See \cite{ayyoubzadeh2020adaptive,elhamod2020cophy,barron2019general,johnson2016perceptual} for just a small sample of recent research.

\subsection{A case in point: using measures from spatial model verification for neural networks}
\label{spatial_verificaton_sec}

Here we briefly discuss one area with particularly high potential for the development and use of custom loss functions, namely neural networks for spatial forecasts in the environmental sciences. 
Meteorologists and other environmental scientists have developed an extensive set of evaluation measures for spatial model verification, and ideally those evaluation measures should be used directly for neural network training for forecast models.  {\it Why train a neural network on anything other than the criteria we seek to optimize?}  However, many networks are still trained using pixel-based measures, such as MAE, MSE, or RMSE, mainly because those are easily available as loss functions.

Gilleland et al.\ \cite{gilleland2009intercomparison} conducted a large scale comparison of different model verification methods that focus on methods to compare a spatial forecast (image) to an observation (also an image). 
In \cite{gilleland2009intercomparison,gilleland2010verifying} they distinguish four primary classes of methods for model verification\footnote{A later classification by the same group cites five \cite{dorninger2018setup}, but we prefer the separation into the original four categories described in \cite{gilleland2009intercomparison,gilleland2010verifying}.}:
\begin{itemize}
    \item 
        {\bf Neighborhood:}
        Methods that apply some kind of neighborhood averaging to both the forecast and the observation before applying a pixel-based comparison of the resulting smoothed images.\\ 
        Example:  Fractions skill score.
    \item 
        {\bf Scale separation:}
        Methods that separate the signals in forecast and observation images into different spatial scales (orthogonal decomposition).\\
        Example: Applying Fourier or wavelet transformation, then spatial filtering, to both images before pixel-wise comparison.
    \item 
        {\bf Features based:}
        Methods that seek to identify features, such as connected regions, in both forecast and observation, then seek to match those features between forecast and observation and compare their properties.\\
        Example: Applying the {\it Method for Object-Based Diagnostic Evaluation (MODE)} framework developed by \cite{brown2007application}, which utilizes fuzzy logic. 
    \item 
        {\bf Field deformation:}
        Method that morphs one of the fields so that its locations match the other field, by producing a field of distortion vectors.\\
        Example:  Using the technique of optical flow to calculate the distortion field.
\end{itemize}
Many approaches fall within more than one category above \cite{gilleland2009intercomparison,gilleland2010verifying}. 
This list indicates the large variety of sophisticated evaluation methods developed by meteorologists over decades.  In contrast, neural networks developed for meteorological forecasting are lagging far behind this development.  In fact, most of them still use simplistic, pixel-based loss functions, to describe the optimal behavior to be achieved by the neural network. Thus there is huge potential to develop more meaningful loss functions that truly optimize the measures that scientists care about.

In particular, one may wonder which ones of the four spatial model verification categories listed above can be implemented in loss functions. 
Both neighborhood based and scale separation based methods are feasible for neural network implementation - in fact we have already implement some of each.  Field deformation methods may become feasible, given that some optical flow algorithms are being implemented as neural networks and that loss functions can call other neural networks without a problem.  We believe the feasibility of that approach depends on how fast, reliable, and accurate those implementations can be in meteorological applications.
Current features based methods are likely the hardest to implement in loss functions, due to their extreme discontinuities when extracting features.  Thus we expect at most some very simple approximations of feature based methods to potentially become feasible for loss function implementation.

\subsection{Organization of this document}

The remainder of this document is organized as follows.
\\[0.2cm]
    {\bf Section \ref{About_this_guide_sec}} provides information on this guide, its intended audience, and the coding environment used for the examples.
\\[0.2cm]
    {\bf Section \ref{basics_sec}} covers introductory material that can be easily found through online resources and in literature. 
    In contrast, the material covered from Section \ref{pitfalls_sec} on goes beyond what is easily accessible through other sources. 
\\[0.2cm]
    {\bf Section \ref{pitfalls_sec}} looks into common implementation pitfalls.  
\\[0.2cm]
    {\bf Section \ref{feeding_more_variables_sec}} discusses how to feed additional information into the loss functions, such as parameters or supplementary information about each sample. 
\\[0.2cm]
    {\bf Section \ref{types_of_functions_sec}} dives into which kinds of functions can be used in loss functions, including unusual but powerful examples, such as how to use existing neural network layer functionality in loss functions.  It also discusses how to properly include conditions in loss functions.
\\[0.2cm]
    {\bf Section \ref{loss_functions_examples_sec}} uses all of these concepts to generate several loss functions for environmental science applications, including loss functions for semantic segmentation (IOU, Dice, Tversky coefficients) and an implementation of the fractions skill score.
\\[0.2cm]
    {\bf Section \ref{advanced_topics_sec}} discusses important loss concepts that have not yet been deeply explored, but that we believe deserve more attention, including incorporating physical constraints, focal loss, robust loss, adaptive loss, the discriminator in generative adversarial networks, and perceptual loss.
\\[0.2cm]
    {\bf Section \ref{vignettes_sec}} provides insights and practical examples from practitioners in environmental science.
\\[0.2cm]
    {\bf Section \ref{conclusions_sec}} provides some concluding comments.

\section{About this guide and its intended audience}
\label{About_this_guide_sec}

This guide was developed by scientists who work at, or collaborate closely with, the Cooperative Institute for Research in the Atmosphere (CIRA) at Colorado State University.  It is simply a collection of the lessons we learned trying to implement meaningful custom loss function for our own applications in the environmental sciences.  
When writing this document we asked ourselves the following question: ``What knowledge do we have now that we wish we had when we started developing custom loss functions?''
This document is thus not written from a computer science perspective, but from a practitioner's perspective.  Namely, it is written {\it by} scientists working in the environmental sciences {\it for} scientists working in the environmental sciences. 

{\bf Disclaimer:}  The contents herein represent our understanding of loss functions to the best of our abilities.  However, there are bound to be some (hopefully minor) mistakes somewhere.  We thus offer the contents, including the code snippets, ``as is,'' without warranty, and disclaim liability for damages resulting from its use.

\subsection{\bf How you can contribute to this effort}
There are several ways in which you, the reader, can help and contribute:
\begin{itemize}
\item 
    If you find any errors (there are bound to be some) or have any other suggestions for improvements, {\it please} let us know.  We intend to fix any bugs in future versions of this document to be posted here and would mention you for bug fixes, etc.  Please email the first author with any comments you may have.
\item
    Do you have loss functions to contribute?  We would be happy to add them to the Github repository (with acknowledgements) and, if of interest, to this guide.
\item
    Are you a PyTorch user and willing to translate some of the examples over to PyTorch?  We would be happy to include those in the Github repository - with proper acknowledgement of course.
\end{itemize} 

{\bf Please cite us:}
If you find this guide useful for your work, {\it please cite it}.  Citing it will help us get recognized for our effort and help us stay motivated to post updated versions from which the entire community may benefit.  With your help, hopefully, we can move the entire community forward on this important topic.

\subsection{\bf Coding Environment}
All discussions and examples here are written based on using Python and Keras with TensorFlow 2 backend. 
TensorFlow/Keras and PyTorch appear to be the most common frameworks used in the meteorological community at the time this manuscript is being written, and it should not be very difficult to transfer the examples provided here in TensorFlow over to PyTorch, as the basic concepts are the same.
Namely, what is possible to code in TensorFlow/Keras should be possible to code in PyTorch in a similar way.  On the other hand, many of the pitfalls described here might be unique to TensorFlow, so are less likely to transfer.

When using Keras we highly encourage the use of the {\it functional} application programming interface (API) to define neural networks (see \url{https://www.tensorflow.org/guide/keras/functional}). 
In contrast to the sequential (old style) API in Keras, the functional API in Keras provides much more flexibility, including for the definition of custom loss functions.  This document assumes that the functional API is used, although many examples will also work with the sequential API. 

TensorFlow functions usually appear as \texttt{tf.function\_name} in the code.  It is always assumed that \texttt{tf} is defined accordingly before.  The equivalent holds for Keras functions, which appear as \texttt{K.function\_name}.  This can be achieved as follows:
\begin{mycodebox}
   import tensorflow as tf
   import tensorflow.keras.backend as K
\end{mycodebox}

{\bf Coding Styles:}
Different examples provided here come from different contributors, which may have different styles of coding.  There are always many ways to do the same thing in Python/TensorFlow/Keras.  We did not try to make the styles uniform, because style is a question of taste and we think it is helpful to see different styles to choose from.  

{\bf Color blocks:}
Throughout this document code samples are indicated by a gray color block in the background.  Pitfalls and warnings are indicated by a cyan color block in the background.

{\bf GitHub repository:}  All code samples included here are available in a GitHub repository at \url{https://github.com/CIRA-ML/custom_loss_functions}.

\subsection{\bf Commonly used acronyms and expressions}
\hspace*{0.0cm}{\bf AI:} Artificial intelligence
\\[0.2cm]
\hspace*{0.0cm}{\bf ML:} Machine learning
\\[0.2cm]
\hspace*{0.0cm}{\bf NN:} Neural network
\\[0.2cm]
\hspace*{0.0cm}{\bf CNN:} Convolutional Neural network
\\[0.2cm]
\hspace*{0.0cm}{\bf NN parameters:} The model parameters of a neural network that are learned during training, namely the weights and biases of all layers. 
\\[0.2cm]
\hspace*{0.0cm}{\bf GOES:} Geostationary Operational Environmental Satellite, see \cite{schmit2017closer} for more information.

\subsection{Other resources}

The TensorFlow and Keras online documentation provide key information on custom loss functions and pointers to many relevant parts of that documentation are provided throughout this document. 
In terms of books, we recommend Geron's machine learning book \cite{geron2019hands}, Chapter 12 of which discusses custom elements for TensorFlow models, including custom metrics, custom loss functions, custom layers, custom activation functions, and much more.  Finally, web forums remain a crucial source of information to find solutions for specific problems that others already discovered.

\section{Introductory material}
\label{basics_sec}

This section provides some introductory material, such as the general purpose of loss functions, metrics, and lambda layers, and introductory material specific to custom loss functions, such as first simple examples and how to save and load a model that contains a custom loss function.
The type of material covered in this section can easily be found in many online resources and in literature, and is included here for convenience.  Many readers can probably skip this section.

\subsection{Custom elements:  metrics, loss functions, and lambda layers}
\label{custom_elements_sec}

Neural network environments, such as TensorFlow and PyTorch, provide many ready-to-use elements, such as standard loss functions, standard metrics, and standard neural network layers (\textit{e.g.,} convolutional and pooling layers).  While for many applications these standard elements are sufficient, we often find that for environmental science applications it is useful to go beyond these fixed elements.  
Using custom elements allows us to do so.  Before diving into these custom elements, here are a few words about the role of these custom elements for neural network development.

{\bf Loss functions:}
The loss function describes what exactly we want the neural network to minimize.  Choosing a loss function is thus critical to truly optimize what we care about the most. Significant time and effort should be dedicated to defining what is most important to the problem and to choose a corresponding loss function.  For example, to compare an observed and estimated satellite image, is a pixelwise comparison using MSE truly the best option?  In some applications that might be sufficient, but in many cases that is sub-optimal.  Furthermore, loss functions, because they are being optimized through gradient descent during the training, must be differentiable.  They must also execute extremely fast to allow for fast training of the neural network.  

{\bf Metrics:}
While we can only define a single loss function, we can use as many metrics as we please while training the neural network.  The metrics are not optimized, but they allow us to track other criteria during and after the training process to spot problems or weaknesses of the trained network early on.  For example, for a binary classification problem one might choose cross-entropy as the loss function, then simultaneously track forecast verification metrics \textendash \, such as hits, misses, false alarms, and correct negatives \textendash \, as auxiliary metrics during training.  These metrics allow to track whether and when the neural network learns to perform well for rare events.  (We will see later that we can also minimize hits, etc., directly in the loss function if we care to do so.)  

Another common scenario is that we might choose a loss function that is a sum of several different criteria, \textit{e.g.,} one criterion focusing on general performance of the network across all samples and another focusing on performance specifically for rare events.  Whenever we use a sum in the loss function, we want to know how much each part of the sum contributes to the loss function, so in this case we recommend to assign a metric for {\it each part} of the loss function - in the example above that would be one for general performance, one for rare events.  Furthermore, whenever L\textsubscript{1} or L\textsubscript{2} regularization is used in TensorFlow layers, the L\textsubscript{1} or L\textsubscript{2} penalty value is automatically added to the value of the loss function itself, and the sum of those values is returned as a combined loss value. To disentangle these penalties from the rest of the loss, we recommend to define a metric that is identical to the loss function, so that it is easy to determine how much of the reported loss is due to the actual loss function (value returned by the metric) and how much is due to the regularization penalty (combined loss value {\it minus} the metric value).

In summary, we highly recommend making ample use of the ability to track any number of metrics simultaneously to learn as much as possible about the performance of the network from many different viewpoints, \textit{i.e.,} tracking a broad selection of performance measures that each focuses on a different type of performance.  Furthermore, not only can we have as many metrics as we want, but they need not be differentiable, as they are not part of the backpropagation routine. Thus, their implementation is much easier, and a much larger variety of performance measures can be used as metrics than as loss functions.
Furthermore, speed is not {\it quite} as crucial, as they are not evaluated as often during training as loss functions.

{\bf Lambda layers:}
Lambda layers provide the ability to create neural network layers that implement a customized transformation from the input neurons to the output neurons of the layer, prescribed as a function.  Just like loss functions, functions in Lambda layers must be differentiable to allow for backpropagation, and they must execute very quickly to avoid slowing down the training process.  In the context of custom loss functions and metrics, lambda layers can be useful to calculate certain quantities that can subsequently be inspected (particularly useful if their output has physical meaning) and used in metrics and loss functions alike. Lambda layers can thus be helpful to calculate intermediate (auxiliary) results in a single place and can thus improve the overall readability and interpretability of code.  Lambda layers are most often used for this purpose as one of the final layers of a neural network, \textit{i.e.,} close to the calculation of the loss and metrics.

\subsection{How evaluation measures can be used in NNs}

\begin{figure}[t]
    \centering
    \includegraphics[width=15cm]{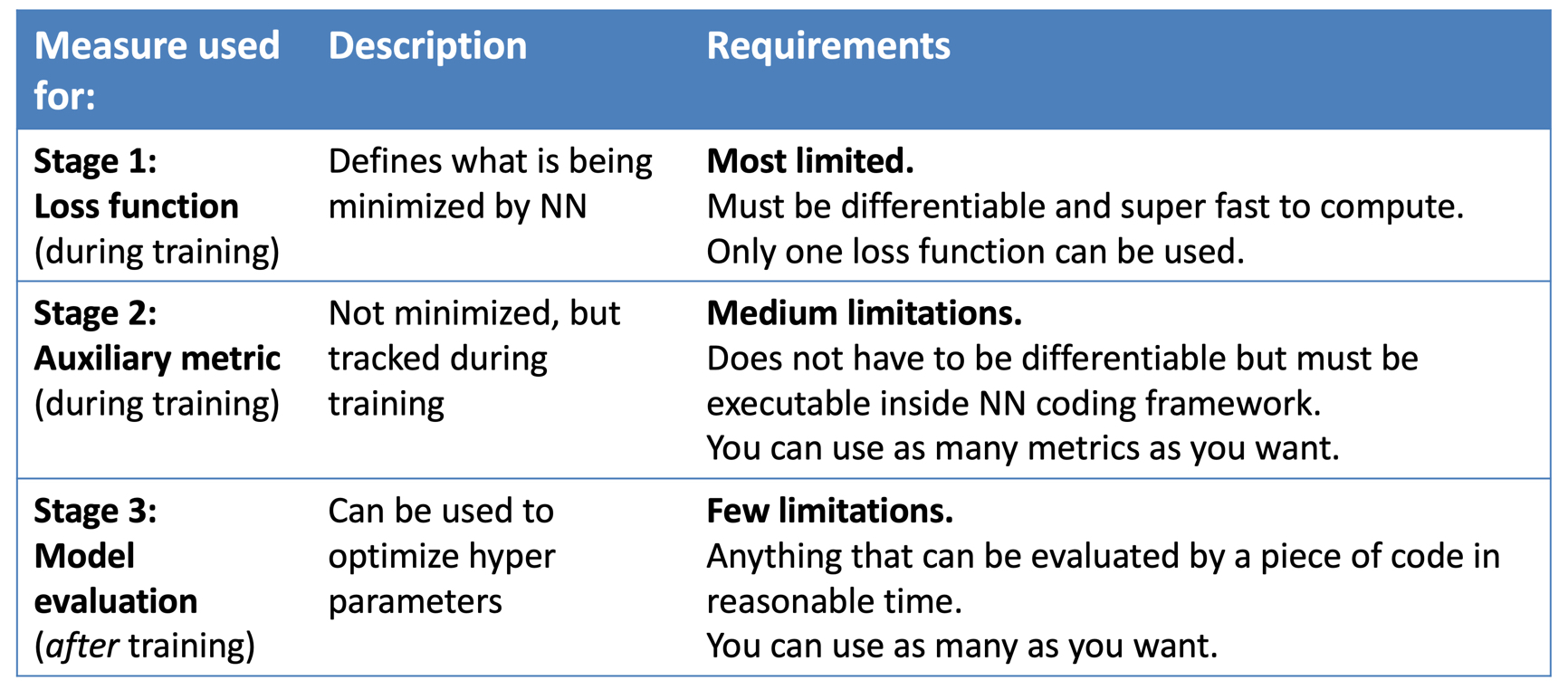}
    \caption{Different stages at which model evaluation measures can be used.}
    \label{NN_evaluation_stages_fig}
\end{figure}

When developing a neural network, existing model evaluation measures can be used at different stages, as shown in Fig.\ \ref{NN_evaluation_stages_fig}.
Optimally, we would like to use evaluation measures in Stage 1, namely as the loss function to be optimized by the neural network.  However, as we already know, loss functions have many restrictions,
thus limiting the use of many desirable model evaluation measures at that stage.
Metrics have fewer limitations, so the range of evaluation measures that can be {\it tracked} by metrics during training is much larger than those that can be {\it optimized} as loss functions.  Finally, for measures that are too complex even to be used as metrics, they can still be used in Stage 3, to evaluate the final model.  Often, they can also still be used to optimize hyperparameters of the NN in an external wrapper, i.e.\ trying different hyperparameters and choosing the model that yields the best results. 
The goal of this write-up is to help the community to move the use of existing model evaluation measures up from Stage 2 (or even Stage 3) to Stage 1, wherever possible.


\remove{
\subsection{Resources}
\label{resources_sec}

1) Keras/TensorFlow documentation:

Metric functions in Keras (existing and custom):\\
\url{https://keras.io/api/metrics/}

Loss functions in Keras (existing and custom):\\
\url{https://keras.io/api/losses/}

Loss functions and metrics in TensorFlow:\\
\url{https://www.tensorflow.org/guide/keras/train_and_evaluate}

2) Books: 
Chapter 12 of \cite{geron2019hands}
} 

\subsection{Defining a custom loss function - first examples}
\label{first_examples_sec}

Here we provide some first example of custom loss functions.

{\bf Defining MSE as custom loss function}

We start with a simple example, namely implementing MSE as a custom loss function.
{\small
\begin{mycodebox}
   def loss_MSE(y_true, y_pred):
      return tf.math.reduce_mean(tf.square(y_true - y_pred))
\end{mycodebox}
}
Items to note: 
\begin{itemize}
\item
    The inputs to the loss function are always two tensors, \texttt{y\_true} and \texttt{y\_pred}, in that order.  \texttt{y\_true} represents the correct output (sometimes called the ``label'' or ``ground truth''), while \texttt{y\_pred} represents the prediction generated by the neural network for that sample.
    You can choose your own variable names for the input tensors, but always make sure that the label (ground truth) is the first input and has a telling variable name, in order to avoid confusion.
\item
    Note that \texttt{y\_true} and \texttt{y\_pred} represent the tensors for an entire batch!  
    We discuss the implications of this fact in detail in Section \ref{batch_considerations_sec}.
\item
    In the above examples the math operations \texttt{sqrt}, \texttt{square}, and \texttt{reduce\_mean} are all TensorFlow functions.  We discuss the types of functions one can use in loss functions in Section \ref{types_of_functions_sec}.
\end{itemize}

{\bf Linking the loss function to a model}

The loss function is linked to the model using the \texttt{model.compile} call, as shown in the example below.
{
\small
\begin{mycodebox}
    model.compile(optimizer=keras.optimizers.Adam(), loss=loss_MSE, metrics=['accuracy'])
\end{mycodebox}
}
Note that there are {\it no} quotation marks placed around the function name, \texttt{loss\_MSE}, above.  The lack of quotes tells Keras that this is a {\it custom} loss function, rather than a built-in loss function. 
In contrast, to use the built-in loss function for MSE, we would call the corresponding function {\it with} quotes:
{
\small
\begin{mycodebox}
  model.compile(optimizer=keras.optimizers.Adam(), loss='mean_squared_error', metrics=['accuracy'])
\end{mycodebox}
}
The metric assigned above, \texttt{accuracy}, also refers to a built-in function, as it is also called with quotes.

{\bf Custom loss function to help with class imbalance}

The previous example was not very exciting, as MSE is available as a standard loss function anyway.  However, we can now create custom functions that help us with specific circumstances.
For example, let us consider an application, such as predicting rainfall, where the great majority of output values is small and only very few values are large.  Since the small values are much more common, the NN can achieve very high performance without ever getting the high values correctly. In other words, because there are only few samples with high values, with a standard loss function the NN might get away with always predicting low values.

There are many ways to deal with that problem, including creating a more balanced data set. Alternatively, we can address this by a custom loss function that penalizes the NN more whenever it gets high values wrong. For example, we can take the standard MSE function and multiply each individual error term by a weight factor that increases exponentially with the true value.  Here is an example that uses $e^{(5 y_{true})}$ as the weight:
{\small
\begin{mycodebox}
   # Loss function with weights based on amplitude of y_true
   def my_MSE_weighted(y_true,y_pred):
      return K.mean(
          tf.multiply(
              tf.exp(tf.multiply(5.0, y_true)),
              tf.square(tf.subtract(y_pred, y_true))
          )
      )
\end{mycodebox}
}
This loss function assigns different weights based on different amplitudes of \texttt{y\_true}: 
\begin{displaymath}
   loss\_MSE\_weighted(y_{\mbox{\small true}}, y_{\mbox{\small pred}})  
   = \mbox{mean}_{i \in I} 
   \left(
   e^{(5 \; y^i_{\mbox{\small true}})} \;  \cdot \;
   (y^i_{\mbox{\small pred}} -y^i_{\mbox{\small true}})^2
   \right),
\end{displaymath}
This is a very simple custom loss function but can already be quite useful.

\subsection{How to save and load a model that has a custom loss or metric} 
\label{save_and_load_model_subsec}

When saving a NN model, unfortunately, custom metrics and loss functions are not stored in the model file. Thus, it is necessary to supply the custom functions when loading the model.

Furthermore, parameters supplied to the custom functions are not automatically stored, either.  There are two ways to deal with this.  One is a manual solution: keep track of the parameters (\textit{e.g.,} in configuration files) and supply them explicitly after the model is loaded.  The more elegant solution is to embed the loss function in a class, which makes the parameters automatically available after loading. This is not discussed here, but interested readers can find it described starting on page 386 of Geron's machine learning book \cite{geron2019hands}.

Consider a model that was trained with a custom loss function and saved in the usual manner.
\begin{mycodebox}
    model.save('K12.h5')
\end{mycodebox}
This example shows how to load that model, which was trained with the following custom loss function,
\begin{mycodebox}
    import tensorflow as tf
    import tensorflow.keras.backend as K
    
    def my_mean_squared_error_weighted_genexp(weight=(1.0,0.0,0.0)):
        def loss(y_true,y_pred):
            return K.mean(tf.multiply(
                tf.exp(tf.multiply(weight,tf.square(y_true))),
                tf.square(tf.subtract(y_pred,y_true))))
        return loss
\end{mycodebox}
and uses the metric,
\begin{mycodebox}
    def my_r_square_metric(y_true,y_pred):
        ss_res = K.sum(K.square(y_true-y_pred))
        ss_tot = K.sum(K.square(y_true-K.mean(y_true)))
        return (1 - ss_res/(ss_tot + K.epsilon()))
\end{mycodebox}
We define both metric and loss in a file called \texttt{custom\_model\_elements.py}. 
If the model is stored in the file \texttt{K12.h5}, as described above, then you can load the model as follows.
\begin{mycodebox}
    from custom_model_elements import my_r_square_metric
    from custom_model_elements import my_mean_squared_error_weighted_genexp
    from tensorflow.keras.models import load_model
    
    model = load_model('K12.h5',
        custom_objects = {
            'my_r_square_metric': my_r_square_metric,
            'my_mean_squared_error_weighted_genexp': my_mean_squared_error_weighted_genexp,
            'loss': my_mean_squared_error_weighted_genexp()
        }
    )
\end{mycodebox}
Because the loss function weights are not loaded using this approach, a model loaded this way cannot be further trained.  However, it can be used to make predictions.

\section{Looking under the hood and common pitfalls to avoid}
\label{pitfalls_sec}

This section briefly reviews the \texttt{compile} command, then, starting with Subsection \ref{batch_considerations_sec}, dives into practical implementation details many of which we have not found discussed anywhere in literature. Bits and pieces were found in web forums and other parts come from our own experience.

\subsection{\bf The \texttt{compile} command in more detail}

The key purpose of the \texttt{compile} command (\texttt{tf.keras.Model.compile}) is to tell the model 
i) which optimizer, ii) which loss function, and iii) which metrics to use (if any). 
It is worthwhile to review for a moment all the options this command provides. 
The \texttt{compile} command has the following format
\begin{mycodebox}
   Model.compile(
      optimizer='rmsprop',
      loss=None,
      metrics=None,
      loss_weights=None,
      weighted_metrics=None,
      run_eagerly=None,
      steps_per_execution=None,
      **kwargs
   )
\end{mycodebox}
The first three parameters are self-explanatory and are usually the only ones used, as seen in the first examples in Section \ref{first_examples_sec}.  We will only discuss one additional optional parameter here, \texttt{loss\_weights}.  For the remaining ones, see the documentation (\url{https://keras.io/api/models/model_training_apis/}).
\begin{itemize}
\item
    \texttt{metrics}: One can provide a list of metrics in the \texttt{compile} command, which will all be tracked {\it separately}.
\item
    \texttt{loss\_weights}: One can also provide a list of loss function terms. However, as there can only be one loss function, Keras will {\it sum} the values of all the provided loss function terms and use the sum as the actual loss function.
    \texttt{loss\_weights} is an optional parameter that allows you to specify an individual weighting factor for each loss function term. 
\end{itemize}

\subsection{Input tensors of a loss function contain samples of an entire batch (not individual samples!)}
\label{batch_considerations_sec}

\mywarning{
Keras, as a high level interface, tends to hide from the user the fact that the two arguments of the loss function, \texttt{y\_true} and \texttt{y\_pred}, represent an entire batch of samples.  That fact, however, has important implications when creating custom loss functions.
}

While the fact that \texttt{y\_true} and \texttt{y\_pred} contain a batch of samples is often mentioned in online resources, it tends to be as a side note and without discussing its implications, and has tripped many of us at first.  Thus we discuss the implications in great detail here.

Let us revisit the MSE loss function in Section \ref{first_examples_sec} (repeated for convenience below):
{\small
\begin{mycodebox}
   def loss_MSE(y_true, y_pred):
      return tf.math.reduce_mean(tf.square(y_true - y_pred))
\end{mycodebox}
}
\texttt{y\_true} and \texttt{y\_pred} are tensors representing an entire batch of samples. 
Namely, \texttt{y\_true} and \texttt{y\_pred} are each of the size 
\begin{mycodebox}
   [batch_size, :]
\end{mycodebox}
Thus the first dimension of each \texttt{y} represents the batch size, while the remaining ones represent the dimensions of \texttt{y} for a single sample.
(We use \texttt{y} here, for brevity, to stand for both \texttt{y\_true} and \texttt{y\_pred}.)
To highlight the impact of \texttt{y\_true} and \texttt{y\_pred} containing batches of samples, rather than individual samples, let us look at the following custom loss function.
{
\small
\begin{mycodebox}
   def loss_RMSE_by_batch(y_true, y_pred): 
      return tf.sqrt(tf.math.reduce_mean(tf.square(y_true - y_pred))) 
\end{mycodebox}
}
For educational purposes we use this as an opportunity to go through every step of this loss function to demonstrate how {\it exactly} Keras applies element-wise (e.g., addition, subtraction, square) and tensor-wise (e.g., \texttt{reduce\_mean}) operations to these tensors.
\\[0.2cm]
{\bf Step 1:}
First the expression
\begin{mycodebox}
   (y_true - y_pred)
\end{mycodebox}
is calculated, which is equivalent to applying the operation 
\begin{mycodebox}
   tf.math.subtract( y_true, y_pred )
\end{mycodebox}
\texttt{tf.math.subtract} is an element-wise operation, so it is applied to all elements of the two tensors simultaneously, across the entire batch and all dimensions of \texttt{y}.  The result is a single tensor with the same dimensions as \texttt{y} that now contains as elements $(y_{true} - y_{pred})$ for all pixels and spanning all samples of the batch.
 
{\bf Step 2:}
The next operation, 
\begin{mycodebox}
   tf.square 
\end{mycodebox}
is also an element-wise operation, yielding as output also a tensor of the same dimensions as \texttt{y}.  Thus we now have a tensor whose elements are $(y_{true} - y_{pred})^2$ for all pixels and spanning all samples of the batch. So far so good. 

{\bf Step 3:}
Now comes the interesting part, taking the mean.  The operation
\begin{mycodebox}
   tf.math.reduce_mean
\end{mycodebox}
completely flattens the tensor from Step 2 to a one-dimensional vector, then takes the mean of that vector.  The mean is taken here including all elements of the tensor, including {\it across all samples in the batch}, resulting in a {\it single scalar} for the entire batch.  

{\bf Step 4:}
The last operation
\begin{mycodebox}
   tf.sqrt 
\end{mycodebox}
takes the square root of the scalar obtained in Step 3.
Note that the mean is taken across all samples of the batch {\it before} the square root is taken.

{\bf Constructing a sample-wise RMSE:}\\
In contrast, if we instead want to use a sample-wise RMSE as loss, (i.e.\ calculate one scalar RMSE for each sample, then take the average {\it afterwards} of these scalars,) then we need to specify the axes in the \texttt{reduce\_mean} function accordingly.
Specifically, we need to take the mean over all dimensions (axes) except the first one, as the first dimension is the batch dimension.  This can be achieved for tensors of arbitrary dimension, using the code snippet below:

{\small
\begin{mycodebox}
   def loss_RMSE_by_sample(y_true, y_pred):
   
      # Reshape each tensor so that 1st dimension (batch dimension) stays intact, 
      # but all dimensions of an individual sample are flattened to 1D.
      y_true = tf.reshape(y_true, [tf.shape(y_true)[0], -1])
      y_pred = tf.reshape(y_pred, [tf.shape(y_true)[0], -1])
   
      # Now we apply mean only across the 2nd dimension, 
      # i.e. only across elements of each sample
      return K.sqrt( tf.math.reduce_mean(tf.square(y_true - y_pred), axis=[1]))
\end{mycodebox}
}
Note that the result of the \texttt{reduce\_mean} command is now a 1D tensor of length \texttt{batch\_size} that contains for each sample the mean of $(y_{true} - y_{pred})^2$ across only that sample. 
In the next step \texttt{tf.sqrt} applies the element-wise square root to all elements of that 1D tensor.  It only remains to take the mean of the resulting 1D tensor.  Note that we do not have to tell TensorFlow to take that mean, since it does this automatically, as discussed below in Section \ref{automatic_mean_sec}.

{\bf Lessons learned:}
\begin{itemize}
\item 
    The fact that \texttt{y\_true} and \texttt{y\_pred} contain samples of an entire batch, rather than individual samples, makes no difference for element-wise operations (e.g., addition, subtraction, multiplication, square, sqrt) in the loss function.
\item
    For any operation that acts across axes (e.g., mean or sum), one should {\bf always specify the axes, in order to be intentional about whether these operations should be applied only within samples or across the entire batch.}
\end{itemize}

\subsection{Mean applied automatically at the end of a loss function}
\label{automatic_mean_sec}

\mywarning{
Many, but not all, loss functions are defined to yield a scalar as output.  Note, however, that if the output is instead a tensor, then Keras will {\it automatically} convert the tensor to a scalar \textendash \, without giving the user any notice \textendash \, by taking the mean of the final tensor, thus yielding a scalar.
}

While this behavior is not a problem by itself, it can lead to confusion.  In particular, if the loss function was supposed to yield a scalar but does not due to some bug, this might go unnoticed, since no warning is provided. 
Adding an \texttt{assert} command to test whether the output is a scalar can be helpful to avoid this pitfall.

As a positive side effect one can use this behavior to shorten the definition of many loss functions.  For example, for the MSE loss function from above 
{\small
\begin{mycodebox}
   def loss_MSE(y_true, y_pred): 
      return tf.math.reduce_mean(tf.square(y_true - y_pred)) 
\end{mycodebox}
}
we can drop the mean without changing the results, since Keras will automatically take the mean anyway: 
{\small
\begin{mycodebox}
   def loss_MSE(y_true, y_pred): 
      return tf.square(y_true - y_pred)
\end{mycodebox}
}
This effect was already used in the function \texttt{loss\_RMSE\_by\_sample} in the preceding subsection.

\subsection{Stateful vs.\ stateless loss functions}
\label{stateful_stateless_sec}

\mywarning{
Custom loss functions by default are {\it stateless}, while many (all?) built-in loss functions are {\it stateful}.  This can result in slight discrepancies between values of built-in versus custom loss functions.
}

A {\it stateless} loss function does not have memory to store the loss values of individual batches and thus reports only the loss value resulting from the {\it last batch} of an epoch.  In contrast, a {\it stateful} loss function has memory, keeps track of the loss of all batches of an epoch, and reports their {\it average} at the end of the epoch.
This can lead to small discrepancies between the loss values reported by custom vs.\ built-in loss functions. Fortunately, the difference between stateful and stateless loss functions does not impact the training of the NN parameters, which are updated after each batch, regardless of whether the loss function is stateful or stateless.  Thus one should not be alarmed by such discrepancies in reported loss values, as they are usually small and do not affect the training.
It is possible to make custom loss functions stateful, but it is quite complex and, for the reasons listed above, usually not worth the effort.

\subsection{Regularizers are added to loss terms}

This effect is widely known, but we mention it here for completeness.

\mywarning{
As mentioned in Section \ref{custom_elements_sec}, each layer with L\textsubscript{1} or L\textsubscript{2} regularization (if there are any) contributes an additional term to the loss function.  These regularization terms are added to the loss itself, whether the loss function is built-in or custom.  This is true for both the training and validation data.
}

Thus, if L\textsubscript{1} or L\textsubscript{2} regularization is used anywhere in the neural network, we recommend keeping track of the loss itself (without regularization terms) in a separate metric.  Regularization terms are not added to the metrics.

\subsection{How loss functions are used during training}

The following happens during neural network training, one batch at a time:
\begin{enumerate}
\item 
    {\bf Loss calculation per batch:}
    As we already know from the preceding subsection, the loss function is called for an entire batch of samples, with \texttt{y\_true} and \texttt{y\_pred} being tensors that contain the entire batch.  The output is a scalar that represents the loss for the entire batch, as discussed in Sections \ref{batch_considerations_sec} and \ref{automatic_mean_sec}. 
\item
   {\bf Gradient calculation per batch:}
   Along with the loss, Keras calculates the gradient of the loss with respect to all trainable parameters simultaneously for an entire batch of samples.  
\item
   {\bf NN parameter update:}
   Once the gradients for all samples in a batch are obtained, the average gradient across the batch is then used to update the NN parameters. 
\end{enumerate}
Training continues until all batches of the training set have been processed. That finishes one epoch.  The loss value is then recorded in the training history, before training moves on to the next epoch.  As discussed in Section \ref{stateful_stateless_sec}, the recorded value represents the loss value of just the last batch of the epoch for stateless loss functions, and the average across the entire batch for stateful loss functions.

The reason why these operations are performed simultaneously for all samples of a batch is two-fold: 
\begin{itemize}
\item 
    To speed up computations, as Graphics Processing Units (GPUs) are particularly good at performing a large number of identical operations in parallel;
\item
    To make training more robust against outliers.  Namely, since the model parameters are only updated for an entire batch and not after each individual sample, a single sample can never move the NN parameters very far away from the current values.
    In other words, the impact of samples on training is smoothed across an entire batch.
\end{itemize}
As a side note, one should thus carefully consider how to choose the batch size. 
It is well known that batch size affects computation time: if we choose it too small the training becomes slow, because we no longer take advantage of the parallel computation capabilities of the GPUs; if we choose it too large we run out of memory.  {\bf However, as is clear from the discussion above, the batch size also impacts how much an individual sample affects the training.}   
We believe the latter effect deserves more study and discussion for environmental science applications.

\section{Feeding additional variables into the loss function}
\label{feeding_more_variables_sec}

Using only the information contained in \texttt{y\_true} and \texttt{y\_pred} in the loss function is quite limiting.  In this section we discuss two ways to feed additional information into the loss function, namely feeding 1) parameters and 2) sample-specific information.  

\subsection{Using function closure to add parameters to the loss function}

Parameters can be added easily, using the concept of function closure in Python.  This concept is widely discussed in most resources on custom loss functions and included here just for completeness.
Function closure allows one to define nested functions, where the inner function inherits the context of the outer function, including variable values.  We can use this concept to add other variables to the inner function, which can be a custom loss function or custom metric.

{\bf Example of function closure in Python:}
{\small
\begin{mycodebox}
   # Nested functions.
   # Scope of outer function is inherited by inner function.

   # Outer function accepts whatever variables you want to give it (a.k.a. the context).
   # Here just one variable: x.

   def f(x):
  
      # Inner function implicitly INHERITS the context from the outer function.
      # Here: value of x.
      
      def g(y):
         # g is a function of y only, but we can nevertheless use x.
         return x * y
         
      return g

   # Create a function.
   g3 = f(3)
   
   # g3 now represents the function g with value x = 3.
   g3(5)   # What do you think the value is?
\end{mycodebox}
}

{\bf Example of function closure to implement dual-weighted MSE as custom loss function:} 

The dual-weighted MSE (similar to that used in \cite{Lagerquist2021_radiation}) is defined as 
\begin{displaymath}
\frac{1}{N} \sum\limits_{i = 1}^{N} \textrm{ max}(y_{true}^{i}, {y}_{pred}^{i})^{\gamma} [y_{true}^{i} - {y}_{pred}^{i}]^2,
\end{displaymath}
where $N$ is the number of data samples; $y_{i}$ is the true value for the $i$\textsuperscript{th} sample; $\hat{y}_{i}$ is the corresponding prediction; and $\gamma$ is a scalar coefficient.  $\textrm{max}(y_{i}, \hat{y}_{i})^{\gamma}$ is a weight that emphasizes samples with a large true or predicted value.  The dual-weighted MSE is implemented as a loss function below, using two nested functions (\textit{i.e.}, function closure).

{\small
\begin{mycodebox}
    def dual_weighted_mse(gamma_weight):
        def loss(target_tensor, prediction_tensor):
            return K.mean(
                (K.maximum(K.abs(target_tensor), K.abs(prediction_tensor)) ** gamma_weight) *
                (prediction_tensor - target_tensor) ** 2
            )

        return loss
    
    # How to use the loss function (with gamma = 5) when compiling a model:
    model.compile(loss=dual_weighted_mse(gamma_weight=5.), optimizer='adam')
\end{mycodebox}
}

\subsection{How to feed sample-specific information into a loss function}

The function closure principle works well to feed parameters into the loss function that are independent of the specific samples.  But what if you want to utilize in the loss function a supplemental quantity that is {\it specific to a sample} and only available during training?

For example, imagine an algorithm to be used in real time for a remote sensing application that outputs images.  There might be information, such as the reliability of the ground truth, that is available only during training because it requires lengthy analysis that is not feasible during real-time operation. Let us say the analysis provides a mask of where the ground truth, \texttt{y\_true}, is reliable.  If we can feed this mask into the loss function, we can focus the neural network training on regions with high reliability (large weight) and use smaller weights for regions with lower reliability, to avoid a potential negative impact of the unreliable ground truth on the training.

A convenient way to feed supplementary information to a loss function, without requiring it to be available when the model is used later for inference, is to add the supplementary information just to \texttt{y\_true}
\footnote{
A very different scenario is when the supplementary information is to be used as additional predictor also during inference.  In that case the supplementary information is added to the sample's input, \texttt{x}, so that the model can explicitly utilize it.  If the supplementary information is {additionally} to be used {\it explicitly} in the loss function, e.g., for differential weighting, then a skip connection can be added to the model that concatenates the input, \texttt{x}, to the NN output, \texttt{y\_pred}. 
For instructions on how to implement the skip connections, see the examples in the official Keras Developers Guide for the Functional API, available at \url{https://keras.io/guides/functional_api/}.}.
This requires a few adjustments:
\begin{itemize}
\item
   The supplementary information, \texttt{y\_suppl}, must be appended to \texttt{y\_true}.  This can be achieved, for example, by concatenation.
\item 
   In the loss function and any metrics, \texttt{y\_true} must be split into two parts, the original ground truth and the supplementary information, which are then utilized accordingly.  Namely, the original ground truth is compared to \texttt{y\_pred}, while utilizing the supplementary information, e.g., for differential weighting of the loss corresponding to entire samples or to specific regions within a generated image.
\item
   The loss function is now receiving two tensors of different types.  \texttt{y\_pred} has the original size, while \texttt{y\_true} is a tensor of size of \texttt{y\_true} combined with \texttt{y\_suppl}.  Fortunately, the requirement of \texttt{y\_true} and \texttt{y\_pred} to be tensors of the same shape has recently been eliminated in TF.  Thus, current versions have no problem accepting \texttt{y\_true} and \texttt{y\_pred} of different shapes.  
   If you get an error, you likely just need to update to a newer version of TF.
\end{itemize}
By channeling the supplemental information only through \texttt{y\_true} to the loss function, no adjustments are needed after the training to run the model in real time without the supplemental information being available, as \texttt{y\_true} is not used anyway when the model is run.

Below are snippets from sample code, with the full working code available in the GitHub repository.  

{\small
\begin{mycodebox}
   # Define a loss function that handles sample-based supplemental information:
   def loss_mse_weighted_input(weights=(1., 1.)):
     """Mean squared error with sample-specific weights."""

      def loss_custom(y_true2D, y_pred):
         y_true = y_true2D[:,0]
         y_suppl = y_true2D[:,1]
         weightsT = tf.where(y_suppl < 1, weights[0], weights[1])
         
         return K.mean(tf.multiply(
             weightsT,
             tf.square(tf.subtract(y_pred, y_true))
         ))

      return loss_custom
  
   # Define any desired metrics, taking into account the new shape of y_true:
   def mse_ytrue2D(y_true2D, y_pred):
      """Custom Mean Squared Error (Ignore Extra Input)"""
      return K.mean(tf.square(tf.subtract(y_pred, y_true2D[:,0])))
    
   # Use these when compiling the model (using adjusted weights of 0.8 and 1.2):
   model.compile(loss=loss_mse_weighted_input((0.8, 1.2)), metrics=[mse_ytrue2D])
\end{mycodebox}
}

In this sample code, for simplicity we assume that both \texttt{y\_suppl} and \texttt{y\_true} are 1D and contain the same number of samples.  \texttt{y\_true2D} can then easily be obtained using concatenation, resulting in a 2D array containing both.


\section{Types of functions to use in custom loss functions and custom metrics}
\label{types_of_functions_sec}

Since NNs are trained by gradient descent, all expressions in the loss function must be differentiable with respect to NN parameters, and should be very fast to execute.  
Furthermore, to avoid the vanishing-gradient problem, loss functions should also not be overly flat, \textit{i.e.,} their derivatives should not be close to zero across large regions of the NN parameter space.
This section first discusses available functions, including TF/Keras math and logic functions and NN layer functions, then goes into how to implement the functionality of \texttt{if} statements. Although \texttt{if} statements are crucial for many measures, they cannot be used in TF loss functions, because they are not differentiable.

\subsection{Large collection of math-related functions available in Keras/TensorFlow modules and the TensorFlow Probability library}

Some general comments about functions to use:
\begin{itemize}
\item
   Generally, functions used in a loss function should all be from TensorFlow or Keras.  Those are optimized for speed.  
\item
   There are a lot more TF math functions than Keras functions available.  Functions that have the same name in TF and Keras have the same functionality. 
   Keras is just the high level interface that calls the backend TF functions.
   One can use functions from both; in fact one can even mix and match both types in the same line of code.
\item
   If in doubt about which version to use (TF or Keras) for these math functions, we prefer to use the TF versions, since they are closer to the source and generally better documented.
 \item
    Math functions available in \texttt{tf.math} include (for a full list, see  \url{https://www.tensorflow.org/api_docs/python/tf/math/}): 
   \begin{itemize}
   \item 
      Basic arithmetic operators (e.g., addition, subtraction, multiplication, sqrt) and trigonometric functions (e.g., sine, cosine).\\ 
      Note: These functions typically act element-wise, i.e., the operation is applied to each element separately.
   \item
      Reductions and scans (e.g., \texttt{reduce\_mean}, \texttt{reduce\_prod}, \texttt{cumsum}).\\
      Note: These functions act across dimensions of a tensor, so care must be taken to specify axes, as discussed in Section \ref{batch_considerations_sec}.
   \item
      Special math functions (e.g., \texttt{igamma} and \texttt{zeta}), complex number functions (e.g., \texttt{imag}, \texttt{angle}) and segment functions (e.g., \texttt{segment\_sum}).
   \end{itemize}
\item
   In addition to these primary math functions in TensorFlow, there are many other TF modules that provide functions of interest that can be used in loss functions and that are currently completely underutilized in the environmental sciences.
   Modules of interest include:
   \begin{itemize}
   \item
      The {\bf linear algebra module (\texttt{tf.linalg})} provides additional math functions for linear algebra.
      \\
      For an overview, see  \url{https://www.tensorflow.org/api_docs/python/tf/linalg}.
   \item 
      The {\bf image operations module (\texttt{tf.image)}} provides many common image processing functions, such as calculating image gradients or the image similarity measure, SSIM, discussed in Section \ref{SSIM_sec}.
      \\
      For an overview, see  \url{https://www.tensorflow.org/api_docs/python/tf/image}.
   \item
      The {\bf signal processing module (\texttt{tf.signal})} provides additional functions.  Of particular interest for environmental sciences are implementations of the fast Fourier transform (FFT) and its inverse in 1D, 2D, and 3D.  
      (Yes, you can call FFT from a loss function!  See upcoming work by Lagerquist and Ebert-Uphoff \cite{Lagerquist2021_loss_functions} on its use for environmental applications.)
      \\
      For an overview, see \url{https://www.tensorflow.org/api_docs/python/tf/signal}.
   \item
      {\bf Many additional modules} of interest are available in TensorFlow. 
      An up-to-date list of all available modules and functions is always available at \url{https://www.tensorflow.org/api_docs/python/tf}.
   \end{itemize}   
\item
   The relatively new {\bf TensorFlow probability library (tfp)} provides many additional functions that can be used in loss functions, 
   e.g., to evaluate probability distributions.
   \\
   For an overview, see \url{https://www.tensorflow.org/probability}. 
\item 
   Lastly, it is also possible to call {\bf neural network layer functions} (e.g., pooling, convolution) from loss functions. This can be very useful to implement meteorological measures of spatial forecasts, as discussed in detail in Section \ref{NN_layers_sec} and demonstrated in Section \ref{FSS_sec} for the implementation of the fractions skill score.
\end{itemize}

\mywarning{
Using functions from numpy is possible and usually does not yield an error, but this might make training {\it extremely slow}.  So always stick with TensorFlow and Keras functions to avoid this problem, as they are guaranteed to be optimized for NN training.}

\subsection{Using NN layer functionality in loss functions and metrics}
\label{NN_layers_sec}

Neural network layers can simply be seen as predefined functions that define a transformation from the input tensor of a layer to an output tensor.  As such, they can be utilized in loss functions and metrics.
Since they are designed for NN layers, they are guaranteed to be differentiable and highly efficient. 

Using predefined layers is very handy to perform predefined image-processing tasks, such as applying a predefined filter to an image for smoothing or edge detection.  Similarly, pooling layers allow to obtain neighborhood averages, which are used in many neighborhood methods for spatial model verification (Section \ref{spatial_verificaton_sec}).

{\bf Sample use of pooling layer:}\\
The following example shows how to use a predefined pooling layer:
\begin{mycodebox}
   pool_1 = tf.keras.layers.AveragePooling2D(
                pool_size=(2, 2),
                strides=(1, 1),
                padding='valid'
            )
            
   y_true_pooled = pool_1(y_true)
\end{mycodebox}
The first line above defines \texttt{pool\_1} as a pooling function by assigning the functionality of a standard Keras pooling layer, specifically here of an average pooling operation in two dimensions with a 2-by-2 kernel size.
The last line above applies this function to a tensor, \texttt{y\_true}.  This functionality is used in the loss function for the fractions skill score (Section \ref{FSS_sec}) to implement the neighborhood averaging.  

{\bf Sample use of convolutional layer:}\\ 
The following example shows how to use a convolutional filter with pre-defined weights.  These weights implement a 5-by-5 Gaussian smoother with $e$-folding radius of 1 pixel.  The example assumes that both the input and output tensor for the convolutional layer have 3 channels; thus, the weight matrix in the convolutional layer must have dimensions of 5 (spatial rows) $\times$ 5 (spatial columns) $\times$ 3 (input channels) $\times$ 3 (output channels).  The 3 channels could be, for example, those in an RGB image: red, green, and blue.
\begin{mycodebox}
    weight_matrix = numpy.array([
        [0.00296902, 0.01330621, 0.02193823, 0.01330621, 0.00296902],
        [0.01330621, 0.0596343, 0.09832033, 0.0596343, 0.01330621],
        [0.02193823, 0.09832033, 0.16210282, 0.09832033, 0.02193823],
        [0.01330621, 0.0596343, 0.09832033, 0.0596343, 0.01330621],
        [0.00296902, 0.01330621, 0.02193823, 0.01330621, 0.00296902]
    ])
    
    # Expand dimensions from 5 x 5 to 5 x 5 x 3 x 3.
    weight_matrix = numpy.expand_dims(weight_matrix, axis=-1)
    weight_matrix = numpy.repeat(weight_matrix, axis=-1, repeats=3)
    weight_matrix = numpy.expand_dims(weight_matrix, axis=-1)
    weight_matrix = numpy.repeat(weight_matrix, axis=-1, repeats=3)
    
    # Conv layer also needs one bias per input channel.  Make these all zero.
    bias_vector = numpy.array([0, 0, 0], dtype=float)
    
    conv_layer_object = keras.layers.Conv2D(
        filters=3, kernel_size=(5, 5), strides=(1, 1),
        padding='same', data_format='channels_last',
        activation=None, use_bias=True, trainable=False,
        weights=[weight_matrix, bias_vector]
    )
\end{mycodebox}

{\bf Sample use of image functions:}  
This example illustrates the use of the Sobel operator, \texttt{tf.image.sobel\_edges}, which can be used to generate an approximation of the gradient of an image, in a loss function. It combines the MSE loss with a weighted contribution from Sobel edges, which can help produce model predictions with sharper spatial gradients.
\begin{mycodebox}
    def my_mse_with_sobel(weight=0.0):
        def loss(y_true,y_pred):
        
            # This function assumes that both y_true and y_pred have no channel dimension.
            # For example, if the images are 2-D, y_true and y_pred have dimensions of
            # batch_size x num_rows x num_columns.  tf.expand_dims adds the channel
            # dimensions before applying the Sobel operator.
            
            edges = tf.image.sobel_edges(tf.expand_dims(y_pred,-1))
            dy_pred = edges[...,0,0]
            dx_pred = edges[...,0,1]
            
            edges = tf.image.sobel_edges(tf.expand_dims(y_true,-1))
            dy_true = edges[...,0,0]
            dx_true = edges[...,0,1]
            
            return K.mean(
                tf.square(tf.subtract(y_pred,y_true)) +
                weight*tf.square(tf.subtract(dy_pred,dy_true)) +
                weight*tf.square(tf.subtract(dx_pred,dx_true))
            )
        return loss
\end{mycodebox}

\subsection{How to implement \texttt{if} statement functionality when conditioning on {\it model-independent} variables: using \texttt{tf.where}}

So far in this section, we discussed functions that are by nature differentiable.  But what if we need functionality that by definition is not differentiable, such as an \texttt{if} statement?  This subsection and the following one deal with two such cases and two different ways of dealing with this problem.

The key question is, what do you need to condition on?  If you need to condition on anything that depends on model parameters, such as the output of a neural network (prediction or classification result), then you cannot use the approach discussed in this subsection, because, well, that function would not be differentiable.  Instead, look at the soft discretization approach discussed in Subsection \ref{discretization_sec}, which can serve as a replacement for hard discretization.
If, on the other hand, you want to condition on anything that does {\it not} depend on the NN model, such as the input data (including NN inputs and labels), then this function is differentiable with respect to model parameters. 
While classic \texttt{if} statements are not allowed in loss functions, 
TensorFlow offers a replacement, \texttt{tf.where} (see \url{https://www.tensorflow.org/api_docs/python/tf/where} for details).
The basic format for the \texttt{where} function is as follows:
{\small
\begin{mycodebox}
   tf.where(condition, expr1, expr2)
\end{mycodebox}
}
which {\it seeks to emulate} the functionality of the following \texttt{if} statement:
\begin{mycodebox}
   [Pseudo code]
   If condition is True:  return expr1 
   Else:	                 return expr2
\end{mycodebox}
However, be aware that this emulation is not perfect.  The pitfalls are easy to understand by looking at the way TF implements the \texttt{where} function, namely by multiplication of \texttt{expr1} and \texttt{expr2} by \texttt{0}s and \texttt{1}s, depending on whether \texttt{condition} is true.
Thus, if you define
{\small
\begin{mycodebox}
   result = tf.where( condition, expr1, expr2 )
\end{mycodebox}
}
TF will in fact internally represent this as 
\begin{mycodebox}
   [Pseudo code]
   If condition is True:  result = 1 * expr1 + 0 * expr2 
   Else:	                 result = 0 * expr1 + 1 * expr2
\end{mycodebox}
One can interpret this as TF defining two different models; depending on \texttt{condition}, one versus the other is used for backpropagation. Each model is differentiable by itself, and TF just switches between them.  Thus, the \texttt{where} function can be interpreted as telling TF when to use one model versus the other - which explains its name.

\mywarning{
\texttt{where} does not play nicely with \texttt{NaN} or other invalid entries\\
The \texttt{where} emulation works fine as long as \texttt{expr1} and \texttt{expr2} represent well defined (finite) numbers.  However, if one of the two expressions is \texttt{NaN} things go very wrong, even if you specifically test for \texttt{NaN} in the \texttt{condition} of \texttt{where}.}

Let us say we want to use the following line to test for and deal with \texttt{NaN} \footnote{Application: A classic example where this can be a pitfall is dealing with missing  or invalid data.  For example, we might have an image with undefined values, as is common in remote sensing imagery, and we want to deal with it on the fly.  We may want to test all values in a sample image for \texttt{NaN} values, define a mask accordingly, and use it in the loss function to just ignore all locations with \texttt{NaN}.}:
{\small
\begin{mycodebox}
   clean_data = tf.where(tf.is_nan(y), C, y)
\end{mycodebox}
}
where \texttt{C} is some (valid) constant.  
Let us consider the case where \texttt{y} is \texttt{NaN}.  
That results in 
{\small
\begin{verbatim}
   clean_data = (1 * C + 0 * y) = (1 * C + 0 * NaN) = NaN
\end{verbatim}
}
because zero times \texttt{NaN} is still \texttt{NaN}!
Having a \texttt{NaN} as the output of a loss function has major repercussions.  Gradient descent can quickly spread the \texttt{NaN} values across the network during backpropagation, resulting in many NN parameters being assigned \texttt{NaN} values. In other words, one occurrence of \texttt{NaN} can easily destroy all progress made up to that point in NN training.

How can we avoid this?  It's simple: whenever you use \texttt{where(\texttt{condition}, expr1, expr2)} in a loss function, make sure that both \texttt{expr1} and \texttt{expr2} contain only valid numbers. The best approach is to fill missing and invalid data before feeding them into the neural network.  As an additional safeguard, one might want to use the \texttt{assert} command to verify inputs to the \texttt{where} function, which prints a warning if \texttt{expr1} or \texttt{expr2} do not represent valid numbers.  That way any problems are caught right away.

Lessons learned: 
\begin{itemize}
\item
   In order for the \texttt{where} function to be compatible with backpropagation, the \texttt{condition} in the \texttt{where} function must not depend on the NN parameters.  That implies that the \texttt{condition} must not depend on \texttt{y\_pred}. However, the \texttt{condition} is allowed to depend on the data samples, such as the true output, \texttt{y\_true}, as well as the network's input data, and on any other supplemental quantities that are independent of the NN parameters.
\item
    \texttt{where} does not play nicely with \texttt{NaN} or other invalid entries in \texttt{expr1} or \texttt{expr2}, so double-check \texttt{expr1} and \texttt{expr2} before feeding them into \texttt{where}.
\end{itemize}

{\bf Example loss function using \texttt{where} for weighting zero and non-zero values differently in the loss function}\\ 
When training on data with imbalanced labels, e.g., quantities that are zero most of the time, such as precipitation, it can help to apply separate weights to zero and non-zero pixels during training.  Here is an example:
\begin{mycodebox}
    def my_mean_sqaured_error_wtzero(weight=(1.0,1.0)):
        def loss(y_true,y_pred):
            ones_array = tf.ones_like(y_true)
            weights_for_zero = tf.multiply(ones_array,weight[0])
            weights_for_nonzero = tf.multiply(ones_array,weight[1])
            
            weights = tf.where(tf.greater(y_true,0),weights_for_nonzero,weights_for_zero)
            
            return K.mean(tf.multiply(weights,tf.square(tf.subtract(y_pred,y_true))))
        return loss
\end{mycodebox}

\subsection{How to implement \texttt{if} statement functionality when conditioning on {\it model-dependent} variables: using soft discretization or raw confidence scores}
\label{discretization_sec}

While the \texttt{where} function is useful to implement conditions on model-independent variables, what if we want to apply conditions on the {\it predicted} values, \texttt{y\_pred}, in the loss function?

A very common application, and the one we primarily focus on here, is the use of NNs for binary classification tasks, i.e., determining whether a certain event is occurring or not. 
A typical NN model for binary classification has two outputs: a confidence score that the event is occurring, $p_{event}$, and a confidence score for no event occurring, $p_{no \; event} = 1 - p_{event}$.  Since $p_{no \; event}$ can be calculated from $p_{event}$, we only care about $p_{event}$.
Note that although the two confidence scores add up to 1.0 and express confidence, they are not guaranteed to be probabilities in the traditional sense, as they are often not calibrated.  We like to call them {\it pseudoprobabilities}, and how we interpret them \textendash \, either as actual probabilities or as vague indications of confidence \textendash \, depends very much on the application and the way the NN was trained.

Regardless of the exact interpretation of confidence scores, classic categorical performance measures are not built on confidence scores, but binary states, which enable counting of events.
For example, to calculate the number of true positives, we count how often the model predicts an event out of all cases where we know that an event actually occurs. Typically, the label, \texttt{y\_true}, is binary and tells us whether an event occurs.  However, \texttt{y\_pred} is given in the form of a confidence score, $p_{event}$, rather than a binary state, which does not directly allow for counting. 
There are three ways to address this issue:
\begin{enumerate}
\item 
    {\bf Hard discretization:} We apply hard discretization to $p_{event}$, e.g., we map all samples with $p_{event} > 0.5$ to \texttt{1} and the remaining ones to \texttt{0}.  This provides binary states for \texttt{y\_true}, which enable direct calculation of classic categorical performance measures.  
\item
    {\bf Soft discretization:} We apply soft discretization to $p_{event}$, which maps all $p$ values to real-valued numbers that tend to be very close to either \texttt{0} or \texttt{1}. This provides an approximation of binary states for \texttt{y\_true}, which, along with modified definitions of categorical performance measures, allows said measures to be approximated.
\item
    {\bf No discretization:} We do not apply any kind of discretization to $p_{event}$; we utilize its original value, instead, for a probabilistic evaluation.  Here we use the same modified definitions for categorical performance measures as in soft discretization, but the results are more probabilistic.
\end{enumerate}
We discuss these three approaches below.  Sample code for all three approaches is provided, too.

\subsubsection{Hard discretization}
\label{hard_discretization_sec}

Hard discretization implements the exact calculation of categorical performance measures by mapping all values of $p_{event}$ to either \texttt{0} or \texttt{1}. 

{\bf Example of hard discretization applied to \texttt{y\_pred} (can be used in metrics but not in loss functions):}

\begin{mycodebox}
    y_pred_binary = tf.where(y_pred > cutoff, 1.0, 0.0)
\end{mycodebox}

This results in value \texttt{1} if \texttt{y\_pred} is larger than the cutoff (a typical cutoff is 0.5, but any can be chosen) and \texttt{0} otherwise.  While this calculates and tracks progress toward the exact categorical performance measures, hard discretization of \texttt{y\_pred} cannot be used in a loss function, because it is non-differentiable.  Thus we need to use one of the other two approaches for NN training.

\subsubsection{Soft discretization}
\label{soft_discretization_sec}

For the case of binary classification, we can achieve {\it soft} discretization by applying a sigmoid function, $S(x)$, to the confidence score, $p_{event}$.  
Fig.\ 2 shows the sigmoid function, $S(x)= \frac{1}{1 + e^{(-x)}}$, as the red solid curve. The sigmoid function is close to \texttt{0} or \texttt{1} for most values of $x$, with a soft transition from \texttt{0} to \texttt{1} near $x=0$. Also shown are two variations of $S(x)$, namely $S(c \cdot x)$ for $c=5$ and $c=0.5$.  The constant, $c$, allows us to choose the speed of transition from \texttt{0} to \texttt{1}.  For larger values of $c$ (blue dashed curve), the transition is much faster and more closely approximates a step function.
\begin{figure}[ht]
    \centering
    \includegraphics[width=10cm]{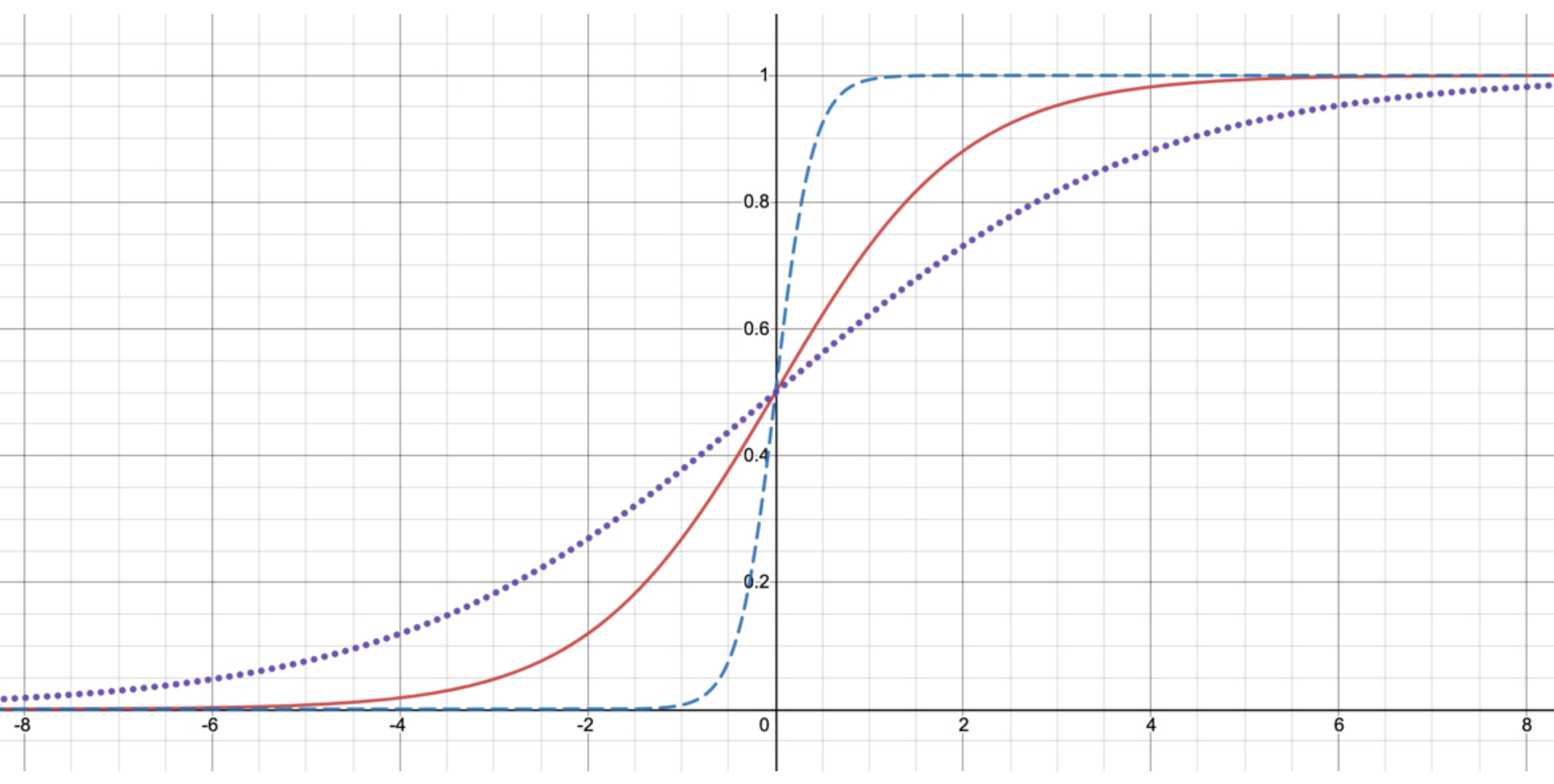}
    \caption{Sigmoid function, $S(x) = \frac{1}{1 + e^{-x}}$ (red solid curve), and two variations thereof, namely $S(c \cdot x)$ for $c=5$ (blue dashed curve) and $c=0.5$ (purple dotted curve). 
    Increasing the value of $c$ results in a faster transition from \texttt{0} to \texttt{1} and closer approximation of a step function. 
    }
    \label{sigmoid_fig}
\end{figure}

Applying the sigmoid function to the confidence score, $p_{event}$, achieves a rough approximation of hard discretization while still being differentiable, so this approach can be used in the loss function.

{\bf Example of soft discretization applied to \texttt{y\_pred} (can be used in both metrics and loss functions):}
\begin{mycodebox}
    c = 5  # constant to tune speed of transition from \texttt{0} to \texttt{1} 
    y_pred_binary_approx = tf.math.sigmoid(c * (y_pred - cutoff))
\end{mycodebox}

Using the formula above results in a value close to \texttt{1} if \texttt{y\_pred} is significantly larger than the cutoff and close to \texttt{0} if \texttt{y\_pred} is significantly smaller than the cutoff \textendash \, \textit{i.e.}, a close approximation of the hard discretization value, \texttt{y\_pred\_binary}, defined in Section \ref{hard_discretization_sec}.  The larger the value of $c$, the closer the soft discretization approximates the step function of hard discretization.

{\bf Choosing the value of the constant, $c$}\\
To choose $c$ we need to remember that loss functions not only have to be differentiable, but they should also avoid being overly flat (small derivative) across large regions of the NN parameter space.  Otherwise we encounter the vanishing gradient problem:  when the gradient diminishes, the update rule in the gradient descent algorithm has no information on how to improve the NN model and learning might stall.
Thus, while soft discretization more closely mimics hard discretization for large values of $c$, this can quickly lead to the vanishing gradient problem, as the sigmoid function is now flat over large regions of $p_{event}$ values.
Smaller values of $c$ lead to more deviation from the hard discretization model, i.e.\ the loss function deviates from the exact measure we want to minimize, but the neural network is likely to converge much faster. 
In other words, we have the choice between (1) having a very precise approximation of the desired measure as a loss function that the neural network has trouble learning, or (2) having a less exact approximation that the neural network can learn much better.  Thus $c$ needs to be chosen as a trade-off.  One might try starting with a default value of $c=1$ and adjust as needed.  In particular, if training stalls one might try a smaller value for $c$. Alternatively, one might include $c$ in a hyperparameter search. During hyperparameter search the NN's performance should be tracked using the exact formula as a metric (using hard discretization), while using soft discretization for optimization in the loss function. 

{\bf Modifying categorical performance measures to accept real-valued numbers}\\
Since $p_{event}$ is mapped to a close approximation of a binary variable but is still real-valued, we need to calculate the counts required for categorical performance measures in a different way. 
Let us say we want to determine the number of samples for which \texttt{y\_pred} is close to \texttt{1}.  The exact value is obtained by first rounding $p_{event}$ for all samples to \texttt{0} or \texttt{1} (i.e.\ applying hard discretization), then counting how many times \texttt{y\_pred} takes the value of \texttt{1}.  The same count can be obtained by summing the rounded values over all samples, 
\begin{displaymath}
   N_{\mbox{event predicted}} = \sum_{\mbox{all samples}} \mbox{Round}( p_{event} ).
\end{displaymath}
Treating the values obtained by soft discretization as an approximation of hard discretization (rounding), we can apply the same formula for the values obtained from soft discretization:  
\begin{displaymath}
   N_{\mbox{event predicted approximation}} = \sum_{\mbox{all samples}} S_c \left( c \cdot (p_{event} - 0.5) \right).
\end{displaymath}
(The cutoff value can, of course, be chosen different from 0.5 if desired.) 

All count-based definitions of categorical performance measures can easily be approximated via this approach.

{\bf Code examples:} 
The hard and soft discretization approaches are both demonstrated in the critical success index in Section \ref{CSI_sec} and in the fractions skill score in Section \ref{FSS_sec}.  

\subsubsection{Using confidence scores to obtain probabilistic measures}

As we discussed at the beginning of Section \ref{discretization_sec}, in some cases we consider the confidence scores as decent proxies of actual probabilities.  In that case we can derive a probabilistic version of the categorical performance measures by simply using the values of $p_{event}$ as is, without any transformation. 
The probabilistic version of the number of events predicted to occur is then obtained by the simple formula,
\begin{displaymath}
   N_{\mbox{event predicted probabilistic}} = \sum_{\mbox{all samples}}  p_{event}.
\end{displaymath}
The motivation for this approach is that the confidence scores actually contain additional information, not contained in the discretized \texttt{0}/\texttt{1} values, which can be useful.  While the confidence scores are not strictly probabilities, they can be treated as proxies for probabilities, and by feeding them into the categorical performance measures, one obtains quasi-probabilistic values instead of deterministic values.
In summary, feeding confidence scores directly into the categorical performance measures can have several advantages:
\begin{itemize}
\item 
   We do not have to tune the constant, $c$, of the sigmoid function;
\item
   We do not increase the risk of vanishing gradients. 
\item
   We can interpret the resulting scores as ``upgraded'' quasi-probabilistic versions of the categorical performance measures, which may actually be a better measure than the corresponding deterministic versions that may have lost significant information through rounding.
\end{itemize}

{\bf Code examples:} This approach is taken in Lagerquist \textit{et al.} \cite{Lagerquist2021_convection} and demonstrated in the CSI example in the following subsection.

\subsubsection{Demonstrating these approaches to implement the critical success index (CSI)}
\label{CSI_sec}

The following example shows how to use the critical success index (CSI) as either a loss function or metric, with any level of discretization (hard, soft, or none).  The CSI is used to evaluate binary classification and is defined as $\frac{a}{a + b + c}$.  The CSI ranges from $\left[ 0, 1 \right]$ and is positively oriented, \textit{i.e.}, higher is better.  Traditionally (assuming hard discretization), $a$ is the number of true positives; $b$ is the number of false positives; and $c$ is the number of false negatives.  Without hard discretization (either soft discretization or none), $a$ is the sum of confidence scores for positive samples; $b$ is the sum of confidence scores for negative samples; and $c$ is the sum of inverted confidence scores (1.0 minus the confidence score) for positive samples.  ``Positive samples'' are those where the event occurs, and ``negative samples'' are those where it does not occur.  

The two input arguments (\texttt{use\_as\_loss\_function} and \texttt{use\_soft\_discretization}) are both Boolean flags, so both may be either \texttt{True} or \texttt{False}.  \texttt{hard\_discretization\_threshold} may be a threshold ranging from $\left[ 0, 1 \right]$ or \texttt{None}; if \texttt{None}, hard discretization will not be used. 
Recall that if \texttt{hard\_discretization} is chosen as \texttt{True} this function can only be used as a metric, not as a loss function. 
\texttt{use\_as\_loss\_function} is necessary because, although CSI is positively oriented, Keras and TensorFlow assume that the loss function is negatively oriented.  If \texttt{use\_as\_loss\_function} is \texttt{True}, the function below returns $(1.0 -\textrm{ CSI})$.  Since the CSI is used only for binary classification, here \texttt{target\_tensor} and \texttt{prediction\_tensor} are both three-dimensional, with dimensions of $E \times M \times N$.  $E$ is the number of examples, while $M$ and $N$ are the number of rows and columns in the spatial grid.  \texttt{target\_tensor} is 1 where the positive class occurs and 0 where the negative class occurs, while values in \texttt{prediction\_tensor} are confidence scores for the positive class.

\begin{mycodebox}
    def csi(use_as_loss_function, use_soft_discretization,
            hard_discretization_threshold=None):
            
        def loss(target_tensor, prediction_tensor):
            if hard_discretization_threshold is not None:
                prediction_tensor = tf.where(
                    prediction_tensor >= hard_discretization_threshold, 1., 0.
                )
            elif use_soft_discretization:
                prediction_tensor = K.sigmoid(prediction_tensor)
        
            num_true_positives = K.sum(target_tensor * prediction_tensor)
            num_false_positives = K.sum((1 - target_tensor) * prediction_tensor)
            num_false_negatives = K.sum(target_tensor * (1 - prediction_tensor))
            
            denominator = (
                num_true_positives + num_false_positives + num_false_negatives +
                K.epsilon()
            )

            csi_value = num_true_positives / denominator
            
            if use_as_loss_function:
                return 1. - csi_value
            else:
                return csi_value
        
        return loss
    
    # How to use the loss function (with no discretization) when compiling a model:
    loss_function = csi(
        use_as_loss_function=True, use_soft_discretization=False,
        hard_discretization_threshold=None
    )
    model.compile(loss=loss_function, optimizer='adam')
\end{mycodebox}


\section{Additional loss functions for environmental science applications}
\label{loss_functions_examples_sec}

This section uses the concepts introduced so far to provide some custom loss functions  
for environmental science applications: 
\begin{itemize}
\item
    Loss functions for semantic segmentation: 
    intersection over union (IOU), Dice coefficient, and Tversky coefficient.
\item 
    A loss function for spatial model verification: fractions skill score (FSS).
\end{itemize}
This section serves to purposes, namely to provide a few ready-to-use loss functions of interest in environmental science and to demonstrate how to use the concepts discussed in the prior sections to create your own loss functions.

\subsection{Loss functions for semantic segmentation}  

This section describes loss functions commonly used for semantic segmentation \cite{Garcia2017}, where classification is performed on each pixel in an image.  Some examples are cancer detection, where the task is to outline cancerous cells in a medical image, and front detection, where the task is to outline fronts in gridded weather data.  Both problems could be either binary or multiclass.  For example, in front detection the task could be to outline warm, cold, stationary, and occluded fronts.  This would be a 5-class problem, as the possible classes also include ``not a front''.  Alternatively, the task could be to outline fronts of any type, which is a binary problem because the only possible answers are ``front'' and ``not a front''.

Another example of semantic segmentation is cyclone detection.  In the work of Kumler \textit{et al.} \cite{kumler2020tropical}, there are 3 possible classes: tropical cyclone, extratropical cyclone, and not a cyclone.  Additionally, Kumler \textit{et al.} experiment with various loss functions for the U-net trained to solve this problem.  They show that while different loss functions lead to broadly similar performance, improvement comes from loss functions that place higher emphasis on the minority classes (tropical and extratropical cyclones) than on the majority class (not a cyclone).  Also, they find that the IOU and Dice coefficient, which both measure the intersection between predicted and actual cyclones, lead to better performance.


We will discuss three loss functions for semantic segmentation: the IOU, Dice coefficient, and Tversky coefficient.  All three loss functions (a) range from $\left[ 0, 1 \right]$ and are positively oriented; (b) can be defined for one class or all classes; (c) can use hard, soft, or no discretization of confidence scores.  For each loss function, to compute the all-class version ($\mathcal{L}$), average the single-class version ($\mathcal{L}_k$) over all classes.  Also, in the equations below, each loss function is defined for only one image pair (predicted and observed).  To compute the single-class or all-class loss for a whole dataset, average $\mathcal{L}_k$ or $\mathcal{L}$, respectively, over all image pairs.  In the code blocks below, both \texttt{target\_tensor} and \texttt{prediction\_tensor} are four-dimensional, with dimensions of $E \times M \times N \times K$.  $E$ is the number of examples (image pairs); $M$ and $N$ are the number of rows and columns in the spatial grid; and $K$ is the number of classes.

{\bf 1) Intersection over Union (IOU):}\\
The IoU for class $k$ is defined as
\begin{equation}
    \textrm{IOU}_k = \frac{\sum\limits_{g = 1}^{G} p_{k}^{g} y_{k}^{g}}{\sum\limits_{g = 1}^{G}\textrm{ max}(p_{k}^{g}, y_{k}^{g})},
    \label{equation:iou}
\end{equation}
where $p_{k}^{g}$ is the confidence score for class $k$ at grid point $g$; $y_{k}^{g}$ is the true label (1 if grid point $g$ belongs to class $k$ and 0 otherwise); and $G$ is the number of grid points in the image.  The numerator is the intersection between the predicted and observed images (where class $k$ both occurs {\it and} is predicted), and the denominator is the union (where class $k$ occurs {\it or} is predicted).  The following example shows how to use the single-class IOU as either a loss function (restricted to \texttt{hard\_discretization\_threshold=None}) or metric.  \texttt{which\_class} is an integer ranging from $\left[ 0, K - 1 \right]$, where $K$ is the number of classes.  If \texttt{which\_class} is $k$, IOU will be computed for the $(k + 1)$\textsuperscript{th} class (since Python uses zero-based indexing).

\begin{mycodebox}
    def iou(use_as_loss_function, use_soft_discretization, which_class,
            hard_discretization_threshold=None):

        def loss(target_tensor, prediction_tensor):
            if hard_discretization_threshold is not None:
                prediction_tensor = tf.where(
                    prediction_tensor >= hard_discretization_threshold, 1., 0.
                )
            elif use_soft_discretization:
                prediction_tensor = K.sigmoid(prediction_tensor)
        
            intersection_tensor = K.sum(
                target_tensor[..., which_class] * prediction_tensor[..., which_class],
                axis=(1, 2)
            )
            union_tensor = (
                K.sum(target_tensor, axis=(1, 2)) +
                K.sum(prediction_tensor, axis=(1, 2)) -
                intersection_tensor
            )

            iou_value = K.mean(
                intersection_tensor / (union_tensor + K.epsilon())
            )

            if use_as_loss_function:
                return 1. - iou_value
            else:
                return iou_value
        
        return loss
    
    # How to use the loss function (with no discretization, for class 1)
    # when compiling a model:
    loss_function = iou(
        use_as_loss_function=True, use_soft_discretization=False, which_class=1,
        hard_discretization_threshold=None
    )
    model.compile(loss=loss_function, optimizer='adam')
\end{mycodebox}

{\bf 2) The Dice coefficient:}
The Dice coefficient for class $k$ is defined as
\begin{equation}
    \textrm{DC}_k = \frac{\sum\limits_{g = 1}^{G} p_{k}^{g} y_{k}^{g}}{G},
    \label{equation:dice_coeff}
\end{equation}
with all variables defined as in Equation \ref{equation:iou}.  The Dice coefficient is similar to the IOU, but the denominator is the area of the full domain ($G$), rather than the intersection for class $k$.  If class $k$ is rare, the Dice coefficient is typically very small, because the full domain is much larger than the intersection for class $k$ (\textit{i.e.}, class $k$ is predicted or observed at very few grid points).  The following example shows how to use the single-class Dice coefficient as either a loss function (again, restricted to \texttt{hard\_discretization\_threshold=None}) or metric.

\begin{mycodebox}
    def dice_coeff(use_as_loss_function, use_soft_discretization, which_class,
                   hard_discretization_threshold=None):
    
        def loss(target_tensor, prediction_tensor):
            if hard_discretization_threshold is not None:
                prediction_tensor = tf.where(
                    prediction_tensor >= hard_discretization_threshold, 1., 0.
                )
            elif use_soft_discretization:
                prediction_tensor = K.sigmoid(prediction_tensor)
        
            intersection_tensor = K.sum(
                target_tensor[..., which_class] * prediction_tensor[..., which_class],
                axis=(1, 2)
            )
            
            # Replacing prediction_tensor with target_tensor here would work, since
            # they both have the same size.
            num_pixels_tensor = K.sum(K.ones_like(prediction_tensor), axis=(1, 2, 3))
            dice_value = K.mean(intersection_tensor / num_pixels_tensor)

            if use_as_loss_function:
                return 1. - dice_value
            else:
                return dice_value
        
        return loss
\end{mycodebox}

{\bf 3) The Tversky coefficient:}\\\
The Tversky coefficient for class $k$ is defined as
\begin{equation}
    \textrm{TC}_k = \frac{\sum\limits_{g = 1}^{G} p_{k}^{g} y_{k}^{g}}{\sum\limits_{g = 1}^{G} \left[ p_{k}^{g} y_{k}^{g} + \alpha p_{k}^{g} (1 - y_{k}^{g}) + \beta (1 - p_{k}^{g}) y_{k}^{g} \right]},
    \label{equation:tversky_coeff}
\end{equation}
where $\alpha$ is the weight for penalizing false positives and $\beta$ is the weight for penalizing false negatives.  Both of these weights are user-selected.  $\sum\limits_{g = 1}^{G} p_{k}^{g} y_{k}^{g}$ is the intersection, where class $k$ both occurs and is predicted, and can be thought of as the number of true positives.  Similarly, $\sum\limits_{g = 1}^{G} p_{k}^{g} (1 - y_{k}^{g})$ is the number of false positives, and $\sum\limits_{g = 1}^{G} (1 - p_{k}^{g}) y_{k}^{g}$ is the number of false negatives.  Thus, in contingency-table terms, Equation \ref{equation:tversky_coeff} could be written as $\frac{a}{a + \alpha b + \beta c}$.  The CSI is $\frac{a}{a + b + c}$, so the Tversky coefficient is essentially a weighted CSI for semantic segmentation.  The following example shows how to use the single-class Tversky coefficient as either a loss function (as always, restricted to \texttt{hard\_discretization\_threshold=None}) or metric.
\begin{mycodebox}
    def tversky_coeff(use_as_loss_function, use_soft_discretization, which_class,
                      false_positive_weight, false_negative_weight,
                      hard_discretization_threshold=None):
                      
        def loss(target_tensor, prediction_tensor):
            if hard_discretization_threshold is not None:
                prediction_tensor = tf.where(
                    prediction_tensor >= hard_discretization_threshold, 1., 0.
                )
            elif use_soft_discretization:
                prediction_tensor = K.sigmoid(prediction_tensor)
        
            intersection_tensor = K.sum(
                target_tensor[..., which_class] * prediction_tensor[..., which_class],
                axis=(1, 2)
            )
            false_positive_tensor = K.sum(
                (1 - target_tensor[..., which_class]) * prediction_tensor[..., which_class],
                axis=(1, 2)
            )
            false_negative_tensor = K.sum(
                target_tensor[..., which_class] * (1 - prediction_tensor[..., which_class]),
                axis=(1, 2)
            )
            denominator_tensor = (
                intersection_tensor + false_positive_tensor + false_negative_tensor +
                K.epsilon()
            )
            tversky_value = K.mean(intersection_tensor / denominator_tensor)

            if use_as_loss_function:
                return 1. - tversky_value
            else:
                return tversky_value
        
        return loss
\end{mycodebox}

\subsection{Fractions skill score (FSS)}
\label{FSS_sec}

Imagine that the output of the neural network is a map estimating precipitation over a certain region (\texttt{y\_pred}), and we also have a map of the actual precipitation (\texttt{y\_true}).  How exactly should the two images be compared in a way that truly measure the similarity of the images?  
A simple (and common) solution is to compare the two images separately for each pixel, e.g., calculating MSE or RMSE across all pixels.
However, that creates the following problem. If strong precipitation is predicted at one location (pixel), but the correct result shows strong precipitation at a neighboring location (pixel), then we get a double error: 
one error for getting precipitation too high at one pixel, another error for missing the high precipitation at the neighboring pixel, although a meteorologist would probably say that our algorithm did a fairly good job. 
Furthermore, the model is penalized as much for this small displacement (one pixel) as for a very large displacement between the predicted and observed precipitation.

\begin{figure}[ht]
    \centering
    \includegraphics[width=12cm]{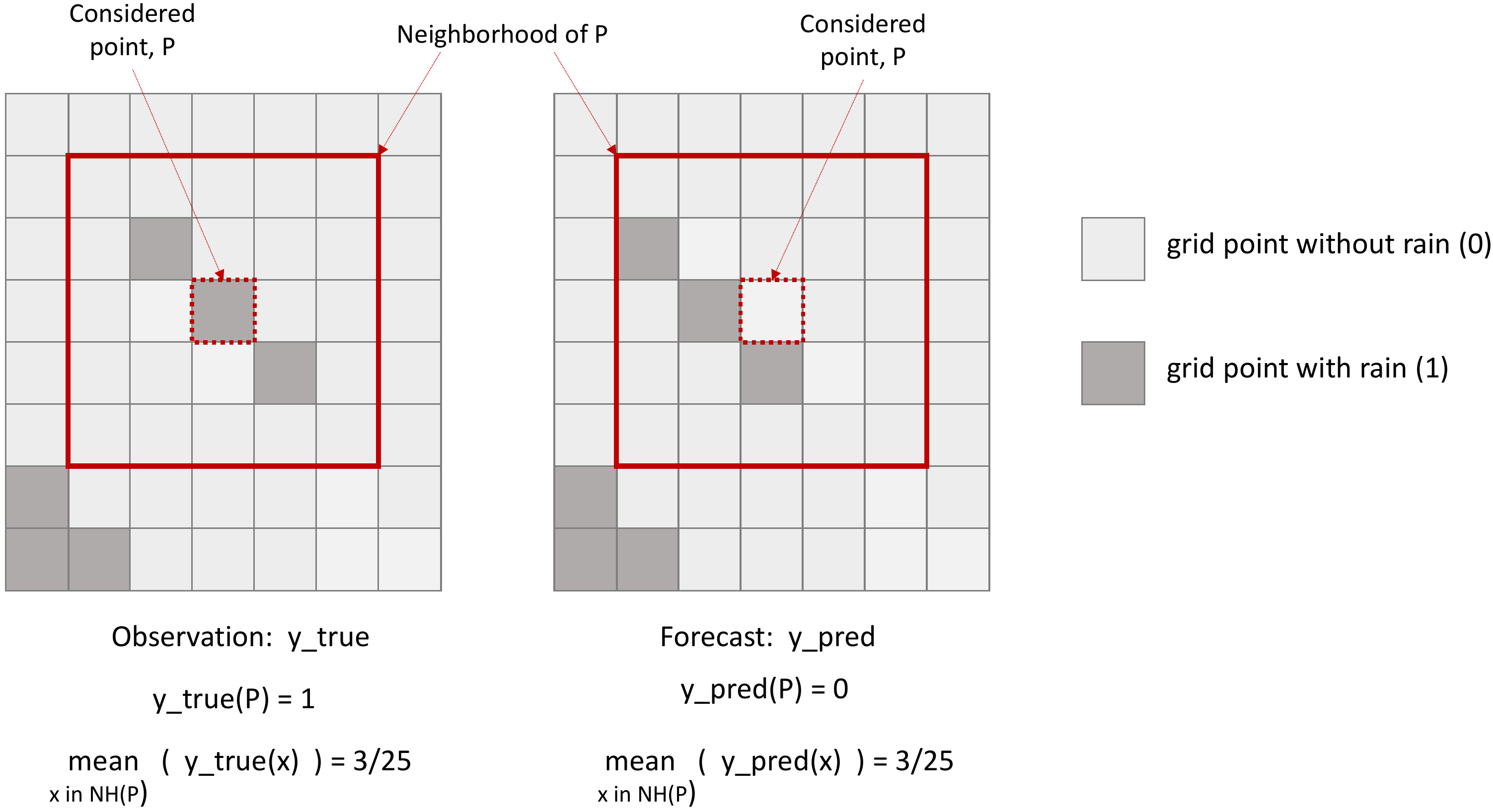}
    \caption{Example of comparing a spatial observation (\texttt{y\_true}) and spatial forecast (\texttt{y\_pred}) pixel by pixel versus by neighborhood. Dark gray squares here represent a grid point with rain, represented by a value of 1, and light gray squares a grid point without rain, represented by a value of 0.
    The forecast looks fairly good overall, but the diagonal rain band is shifted one grid point to the left.  Thus comparing \texttt{y\_true} and \texttt{y\_pred} for this example {pixel by pixel} would result in several classic double counting errors.  For example, at point P the forecast would be penalized for not indicating rain, and one pixel to the left of P, it would be penalized for indicating rain.   
    However, if instead the average is taken over a 5x5 neighborhood centered at P, NH(P), then the value at P is $3/25$ in both images, i.e.\ the forecast would be deemed perfect at P, considering a 5x5 average.}
    \label{FSS_image_from_paper}
\end{figure}

There are several solutions to this problem. 
Here, we focus on the solution proposed by Roberts and Lean \cite{roberts2008scale},
namely the fractions skill score (FSS), and show how to encode it into a loss function.  The FSS is commonly used in traditional meteorological applications to compare a predicted image (forecast) to a correct image (observation) at different scales.  The key idea of the fractions skill score is to compare the correct and estimated images at different resolutions, based on neighborhood averaging. Lower resolution versions of the true and predicted images are obtained from the full resolution images by applying a filter that takes the mean of neighboring pixels, as indicated in Figure \ref{FSS_image_from_paper}. For more details, see \cite{roberts2008scale}.

As discussed in Section \ref{NN_layers_sec}, 
calculating neighborhood averages is exactly what pooling layers in a neural network achieve; thus, we can use a pooling layer in the loss function to easily implement this operation, as shown below. 
Note that an unusual stride size of one is used in the pooling operation to implement the same operation as the neighborhood averaging in \cite{roberts2008scale}.
Traditionally, the FSS in \cite{roberts2008scale} applies only to binary images, so we need to use one of the two strategies discussed in Section  \ref{discretization_sec} for loss functions, namely either apply soft discretization or use the raw confidence scores provided by the neural network.  Either way, the resulting FSS is quasi-probabilistic.

{ 
\footnotesize
\begin{mycodebox}
# Function to calculate "fractions skill score" (FSS).
#
# Function can be used as loss function or metric in neural networks.
#
# Implements FSS formula according to original FSS paper:
#    N.M. Roberts and H.W. Lean, "Scale-Selective Verification of
#    Rainfall Accumulation from High-Resolution Forecasts of Convective Events",
#    Monthly Weather Review, 2008.
# This paper is referred to as [RL08] in the code below.
    
def make_FSS_loss(mask_size):  # choose any mask size for calculating densities

    def my_FSS_loss(y_true, y_pred):

        # First: DISCRETIZE y_true and y_pred to have only binary values 0/1 
        # (or close to those for soft discretization)
        want_hard_discretization = False

        # This example assumes that y_true, y_pred have the shape (None, N, N, 1).
        
        cutoff = 0.5  # choose the cut off value for discretization

        if (want_hard_discretization):
           # Hard discretization:
           # can use that in metric, but not in loss
           y_true_binary = tf.where(y_true>cutoff, 1.0, 0.0)
           y_pred_binary = tf.where(y_pred>cutoff, 1.0, 0.0)

        else:
           # Soft discretization
           c = 10 # make sigmoid function steep
           y_true_binary = tf.math.sigmoid( c * ( y_true - cutoff ))
           y_pred_binary = tf.math.sigmoid( c * ( y_pred - cutoff ))

        # Done with discretization.

        # To calculate densities: apply average pooling to y_true.
        # Result is O(mask_size)(i,j) in Eq. (2) of [RL08].
        # Since we use AveragePooling, this automatically includes the factor 1/n^2 in Eq. (2).
        pool1 = tf.keras.layers.AveragePooling2D(pool_size=(mask_size, mask_size), strides=(1, 1), 
           padding='valid')
        y_true_density = pool1(y_true_binary);
        # Need to know for normalization later how many pixels there are after pooling
        n_density_pixels = tf.cast( (tf.shape(y_true_density)[1] * tf.shape(y_true_density)[2]) , 
           tf.float32 )

        # To calculate densities: apply average pooling to y_pred.
        # Result is M(mask_size)(i,j) in Eq. (3) of [RL08].
        # Since we use AveragePooling, this automatically includes the factor 1/n^2 in Eq. (3).
        pool2 = tf.keras.layers.AveragePooling2D(pool_size=(mask_size, mask_size),
                                                 strides=(1, 1), padding='valid')
        y_pred_density = pool2(y_pred_binary);

        # This calculates MSE(n) in Eq. (5) of [RL08].
        # Since we use MSE function, this automatically includes the factor 1/(Nx*Ny) in Eq. (5).
        MSE_n = tf.keras.losses.MeanSquaredError()(y_true_density, y_pred_density)

        # To calculate MSE_n_ref in Eq. (7) of [RL08] efficiently:
        # multiply each image with itself to get square terms, then sum up those terms.

        # Part 1 - calculate sum( O(n)i,j^2
        # Take y_true_densities as image and multiply image by itself.
        O_n_squared_image = tf.keras.layers.Multiply()([y_true_density, y_true_density])
        # Flatten result, to make it easier to sum over it.
        O_n_squared_vector = tf.keras.layers.Flatten()(O_n_squared_image)
        # Calculate sum over all terms.
        O_n_squared_sum = tf.reduce_sum(O_n_squared_vector)

        # Same for y_pred densitites:
        # Multiply image by itself
        M_n_squared_image = tf.keras.layers.Multiply()([y_pred_density, y_pred_density])
        # Flatten result, to make it easier to sum over it.
        M_n_squared_vector = tf.keras.layers.Flatten()(M_n_squared_image)
        # Calculate sum over all terms.
        M_n_squared_sum = tf.reduce_sum(M_n_squared_vector)
    
        MSE_n_ref = (O_n_squared_sum + M_n_squared_sum) / n_density_pixels
        
        # FSS score according to Eq. (6) of [RL08].
        # FSS = 1 - (MSE_n / MSE_n_ref)

        # FSS is a number between 0 and 1, with maximum of 1 (optimal value).
        # In loss functions: We want to MAXIMIZE FSS (best value is 1), 
        # so return only the last term to minimize.

        # Avoid division by zero if MSE_n_ref == 0
        # MSE_n_ref = 0 only if both input images contain only zeros.
        # In that case both images match exactly, i.e. we should return 0.
        my_epsilon = tf.keras.backend.epsilon()  # this is 10^(-7)

        if (want_hard_discretization):
           if MSE_n_ref == 0:
              return( MSE_n )
           else:
              return( MSE_n / MSE_n_ref )
        else:
           return (MSE_n / (MSE_n_ref + my_epsilon) )

    return my_FSS_loss
\end{mycodebox}
}

\section{Other important concepts for loss functions}
\label{sec:advanced_topics}
\label{advanced_topics_sec}

In this section we discuss additional concepts for loss functions that we believe are very important for environmental science applications, yet have not been extensively explored in that area.

\subsection{Incorporating physical constraints}
\label{subsec:physical_constraints}

One way to add expert knowledge into ML algorithms is to integrate known physical constraints into loss functions.  There are many benefits to doing so \cite{karpatne2017theory,willard2020integrating}, such as generating outputs that are physically consistent, making the most out of a small data set, and generating models that generalize better, i.e.\ perform better on samples from regimes they have not been trained on.
For example, \cite{beucler2021enforcing} shows how constraints can be integrated into neural networks, in either the architecture or loss function.  \cite{willard2020integrating} provides a survey of loss-based approaches to incorporate physics, as well as many other approaches for incorporating physics into machine learning. 

One example is in Lagerquist \textit{et al.} \cite{Lagerquist2021_radiation}, who use a neural network to predict radiative transfer (the propagation of solar radiation through the atmosphere).  Two of their output variables are surface downwelling flux ($F_\downarrow$) and top-of-atmosphere upwelling flux ($F_\uparrow$).  Net flux ($F_{\textrm{net}}$) is related to these variables by the following law:
\begin{equation}
    F_{\textrm{net}} = F_\downarrow - F_{\uparrow}.
    \label{equation:net_flux}
\end{equation}
In initial experiments, the neural network predicted only $F_\downarrow$ and $F_\uparrow$ explicitly; $F_{\textrm{net}}$ was computed by post-processing, outside of training.  The authors found that despite good predictions of $F_\downarrow$ and $F_\uparrow$, the neural network often made poor predictions of $F_{\textrm{net}}$, due to compounding biases (\textit{e.g.}, $F_\downarrow$ too low and $F_\uparrow$ too high, resulting in $F_{\textrm{net}}$ much too low).  In these initial experiments the loss function was the MSE on $F_\downarrow$ and $F_\uparrow$.  The authors added $F_{\textrm{net}}$ to the loss function while enforcing the law in Equation \ref{equation:net_flux}, resulting in
\begin{equation}
    \begin{split}
        \mathcal{L} &= \frac{1}{N} \sum\limits_{i = 1}^{N} \left[ (F_\downarrow^{(i)} - \hat{F}_\downarrow^{(i)})^2 + (F_\uparrow^{(i)} - \hat{F}_\uparrow^{(i)})^2 + (F_{\textrm{net}}^{(i)} - \hat{F}_{\textrm{net}}^{(i)})^2 \right] \\
        &= \frac{1}{N} \sum\limits_{i = 1}^{N} \left[ (F_\downarrow^{(i)} - \hat{F}_\downarrow^{(i)})^2 + (F_\uparrow^{(i)} - \hat{F}_\uparrow^{(i)})^2 + (F_\downarrow^{(i)} - F_\uparrow^{(i)} - \hat{F}_\downarrow^{(i)} + \hat{F}_\uparrow^{(i)})^2 \right].
    \end{split}
    \label{equation:net_flux_loss}
\end{equation}
$N$ is the number of data examples; $F_\downarrow^{(i)}$ and $\hat{F}_\downarrow^{(i)}$ are the observed and predicted $F_\downarrow$ values for the $i$\textsuperscript{th} example; and other variables are defined analogously.   Including the physical constraint of Equation \ref{equation:net_flux} in the loss function (Equation \ref{equation:net_flux_loss}) significantly improved predictions of $F_{\textrm{net}}$, with no negative impact on $F_\downarrow$ or $F_\uparrow$.  Specific results are shown in Vignette 3 of Section \ref{sec:recommendations}.


\subsection{Dealing with hard-to-learn samples: focal loss and robust loss}

{\it Focal loss functions} \cite{lin2017focal} seek to address a certain type of class imbalance problem.  
Namely, imagine a task where you need to detect events, where the events are rare and some of them are very hard to detect (e.g., due to a very complex pattern), while all the non-events are easy to classify.  As a result most of the samples are extremely easy to classify: the great majority are non-events that are easy to classify, and even most of the actual events are easy to classify. 
The great majority of easy-to-detect samples can overwhelm the NN training, giving little attention to the few hard-to-solve samples. 
Focal loss seeks to overcome this problem by (1) giving more weight to the rare events and (2) in addition giving {\it even more weight} to the hard-to-classify examples among the rare events.  
Focal loss was originally developed for object detection \cite{lin2017focal}, i.e.\ for distinguishing foreground and background areas in an image, and furthermore looking for a specific object type in the areas identified as foreground.
Lin et al.\ \cite{lin2017focal} show that models with simple (one-stage) architecture that use their focal loss outperform even much more complex (two-stage) architectures that use regular loss functions. The TF module \texttt{tfa} (\url{https://www.tensorflow.org/addons/api_docs/python/tfa}) is a third-party extension of TF and provides an implementation of the loss function used in \cite{lin2017focal}, called
\texttt{SigmoidFocalCrossEntropy}.
\cite{ma2021loss} extends the focal concept to several other loss functions and performs rigorous comparisons for several medical image segmentation benchmarks.

But what if the hard-to-solve cases are for some reason so hard that the considered neural network is unable to solve them at all?  Those cases may be due to data samples with large noise, corrupted data, unreliable ground truth, etc.  In that case the hard-to-solve samples can become detrimental to the NN training, and then the opposite strategy is required, namely to {\it de-emphasize} the hard-to-solve cases in the loss function so that the NN focuses primarily on getting the easy-to-solve cases right, without being derailed by the hard-to-solve cases.  This type of problem responds well to so-called {\it robust loss functions}, which were developed specifically to be robust to noise in samples.
Robust loss functions include the well known {\it Huber loss} \cite{meyer2019alternative}, which makes small adjustments by implementing a middle ground between MAE and MSE. (Huber loss is available as built-in loss function in Keras.)
More sophisticated methods include the {\it adaptive} robust loss function framework described in \cite{barron2019general}.
The work of highest relevance to this community is by Barnes and Barnes \cite{barnes2021classification,barnes2021regression}. They discuss the robust loss method of {\it abstention networks}, with special emphasis on how to apply it to earth science applications.  Not only do they demonstrate the potential of the abstention framework for classification tasks in earth science \cite{barnes2021classification}, they also extend the framework to regression problems \cite{barnes2021regression}, a major step to making this framework applicable to many earth system applications.

\subsection{Adaptive loss functions}

In a standard neural network the loss function is a fixed formula, i.e.\ it does not change during training. However, the same loss function might not always be optimal or effective during training as the model continues to learn and improve. The idea of an {\it adaptive loss function}, therefore, has been raised to include information from the training so that the model can incorporate and adjust to what it has previously learned. There are different ways of applying the adaptive loss function concept for different purposes. 

{\bf Training in discrete phases:}\\
\cite{lee2020applying} seeks to identify convection from satellite imagery and uses two loss function sequentially to train the model.
The overall motivation for this type of approach is that the network first seeks to achieve an overall objective, then, after hopefully reaching the neighborhood of the desired model, it is fine tuned with the second loss function that additionally includes a secondary objective. 
For example, in \cite{lee2020applying} the model is first trained with MSE as the loss function, which penalizes both misses and false alarms, for a fixed number of epochs.  
The number of epochs is chosen (through trial) large enough to ensure the model has converged to a plateau.  After this initial training the model is trained further, using a loss function that consists of MSE plus an additional term that adds an extra penalty specifically when the model prediction misses convection (but not for false alarms), since for the application in \cite{lee2020applying} we care more about preventing misses than false alarms:
\begin{mycodebox}
   # Standard MSE loss function
   def my_MSE_per_pixel( y_true, y_pred ):
      return K.square(y_pred - y_true)
   
   # Standard MSE loss function plus term penalizing only misses 
   def my_MSE_fewer_misses ( y_true, y_pred ):
      return K.square(y_pred - y_true) + K.maximum((y_true - y_pred), 0)
\end{mycodebox}

Such a switch of loss functions is easy to implement.  One simply calls the combination of \texttt{model.compile} and \texttt{model.fit} twice, once with the first loss function, then with the second loss function, both with a fixed number of epochs:
\begin{mycodebox}
   ### Training Phase 1
   # Assign first set of loss functions, optimizer and metrics:
   metric_list = [
       'mean_squared_error', my_count_true_convection, my_count_pred_convection, 
       my_overlap_count, jaccard_distance_loss
   ]
   model.compile(
       loss=my_MSE_per_pixel, metrics=metric_list, optimizer=RMSprop()
   )

   # First training phase (100 epochs from scratch):
   history = model.fit(
       [x_train_vis,x_train_ir], y_train, 
       validation_data=([x_test_vis,x_test_ir], y_test),
       epochs=100, batch_size=10
   )

   ### Training Phase 2
   # Assign second set of loss functions, optimizer and metrics:
   model.compile(
       loss=my_MSE_fewer_misses, metrics=metric_list, optimizer=RMSprop()
   )

   # Second training phase (fine tune model from Phase 1 with new loss function):
   history = model.fit(
       [x_train_vis,x_train_ir], y_train,
       validation_data=([x_test_vis,x_test_ir], y_test),
       epochs=100, batch_size=10
   )
\end{mycodebox}
This approach achieves the desired effect since all that \texttt{model.compile} does is to define the loss function, optimizer, and metrics.  In contrast, \texttt{compile} does {\it not} reinitialize the NN parameters; thus, in Phase 2 this code continues training the NN that was obtained in Phase 1. Note that you lose the optimizer states when calling \texttt{compile} a second time, i.e.\ the learning rate needs to adjust anew. 

More generally, one can use the approach of calling \texttt{model.compile} and \texttt{model.fit} repeatedly with different loss functions, or with different parameters for a loss function, to implement many phases of training. Furthermore, one can use the output from one training phase to automatically choose a parameter (or loss function) for the next training phase.  This multi-phase approach is probably the easiest way to implement adaptive loss functions.

{\bf Continuous adjustment of a parameter inside a loss function:}\\ 
Another option to implement an adaptive loss function is to create a loss function with a parameter that is changed continuously during training. 
For example, Elhamod et al.\ \cite{elhamod2020cophy} consider three physics-based terms in a loss function that are competing against each other. Figure 1 in \cite{elhamod2020cophy} provides a clear motivation to use adaptive loss functions, demonstrating that a fixed loss function would not lead to a global minimum for their example.  They create a single loss function that adds all three loss terms, but with weight factors for each loss term, which are adjusted throughout training.  They demonstrate that this approach is superior and can avoid ending up in a sub-optimal local minimum for their example.

\cite{barron2019general} developed an adaptive loss function that allows a model to automatically learn to distinguish between inliers and outliers, to reduce the effect of outliers in the training set.  The proposed loss function is a generalized loss function that can represent several traditional loss functions, depending on the values of a single parameter. Those traditional loss functions embedded in the generalized loss function have different sensitivity to outliers, and thus, by training the parameter in the generalized loss function, the model can be trained with different sensitivity to outliers to find the best fit. 
This parameter is adjusted through part of the neural network's training. For details, see \cite{barron2019general}.

Finally, in the case of a recurrent network, time information can be incorporated into the loss function, as described by \cite{suzuki2018anticipating}. The goal of that work is to develop a system for early anticipation of traffic accidents, and they use a loss function that penalizes the model differently, depending on its early anticipation skill. 
%


\subsection{The ultimate adaptive loss function: discriminator NN in GANs}

Adaptive loss functions typically consist of one or more families of functions with adjustable parameters.  Adaptation in that case means switching between different families and/or adjusting a parameter within each family. 
Generative adversarial networks (GANs) \cite{goodfellow2014generative}
take the idea of an adaptive loss function to a whole new level - by creating a {\it flexible, self-learning} loss function, known as the discriminator.

A GAN consists of two neural networks.  One is a primary NN, known as the {\it generator}, which generates output to achieve a certain task, e.g., generate high resolution images from low resolution images.  The other is a secondary NN, known as the {\it discriminator}, which judges the quality of the generator's output during the GAN's training phase.  Once training is completed only the generator is used. 
Of particular interest to the environmental sciences are conditional GANs (cGANs), such as Pix2Pix \cite{isola2017image}. Examples of cGANs used in earth sciences include \cite{stengel2020adversarial,wang2018enhancing}.

The discriminator NN can be seen as implementing an extremely versatile  
loss function for the given task - it learns {\it from examples} how to judge the generator's output. 
Namely, the discriminator is given examples of correct and incorrect outputs, along with the label of whether those examples are correct, and extracts desired output properties from those.
{If the discriminator is shown a large enough selection of correct and incorrect sample outputs, it can learn abstract features that the generator output must have, such as specific small scale features in images or even physical properties \cite{stengel2020adversarial}.} Some of those features would be extremely difficult or impossible to derive in explicit mathematical terms, i.e.\ as a human-made loss function.  Thus the power of the discriminator approach comes from it extracting its own evaluation rules, rather than being given an evaluation rule by a prescribed loss function.
Using such a self-learning loss function can be extremely powerful, especially for synthetic image generation.  
The flip side is that the complexity is much higher, training becomes harder, and a self-learning loss function can create yet another level of abstraction and opaqueness, thus making it even harder to understand how the primary neural network, the generator, is being trained.  Nevertheless, \cite{stengel2020adversarial} demonstrates that cGANs can learn and emulate physically meaningful small scale features for downscaling applications in climate science that to date cannot be achieved by neural networks with standard loss functions.  We have yet much to learn about how to make the most of cGANs in earth science applications, including the most suitable applications, how to best train them in a reliable and efficient manner, and how to distinguish when the resulting output (especially images) just {\it look realistic} versus when they are {\it highly accurate}. 

\subsection{Structural similarity index measure (SSIM)}
\label{SSIM_sec}

In addition to the custom loss functions for image comparison discussed so far we discuss here a built-in function in TF that is well known in the remote sensing literature \cite{guo2016image,yang2011optimized}, but not in the meteorological literature.
The {Structural Similarity Index Measure (SSIM)}, developed by Wang et al.\ \cite{wang2002universal}, is one of the most common measures in the computer vision literature to measure the similarity of two images.
It is so common that it is already implemented in TF.  

An excellent, intuitive description of the SSIM can be found in \cite{hore2010image}, which includes the following description: ``[The SSIM] is considered to be correlated with the quality perception of the human visual system (HVS).  Instead of using traditional error summation methods, the SSIM is designed by modeling any image distortion as a combination of three factors that are loss of correlation, luminance distortion and contrast distortion.''  It should be noted that some authors disagree with the notion that the SSIM matches human perception particularly well \cite{dosselmann2011comprehensive}.  See Section \ref{VGG_sec} for more sophisticated measures to match human perception.

The SSIM between two images, $I_1$ and $I_2$, is the product of three terms:
\begin{displaymath}
   SSIM( I_1, I_2 ) = l(I_1, I_2) c(I_1, I_2) s(I_1, I_2),  
\end{displaymath}
where 
\begin{eqnarray*}
   l(I_1,I_2) &=& \frac{2 \mu_1 \mu_2 + C_1}{\mu_1^2 + \mu_2^2 + C1}
   \qquad \mbox{Luminance comparison: maximal and equal to 1 for $\mu_1=\mu_2$.}\\
   c(I_1,I_2) &=& \frac{2 \sigma_1 \sigma_2 + C_2}{\sigma_1^2 + \sigma_2^2 + C_2} \qquad \mbox{Contrast comparison: maximal and equal to 1 for $\sigma_1=\sigma_2$.}\\
   s(I_1,I_2) &=& \frac{2 \sigma_{12} + C_3}{\sigma_1 \sigma_2 + C_3} 
   \qquad \mbox{Structure comparison (correlation($I_1$,$I_2$)): maximal and equal to 1 for $I_1=I_2$.}
\end{eqnarray*}
The $C_i$ are small constants to avoid a null demonimator;
$\mu_i$ is the local mean of image $I_i$ at a specific pixel (local mean luminance);
$\sigma_i$ is the local standard deviation of image $I_i$ at a specific pixel (local contrast of each image); 
and $\sigma_{ij}$ is the covariance between $I_1$ and $I_2$.

The user chooses the size of the area over which the local mean is to be taken, typically in the form of a Gaussian filter defined by a size and a width parameter. 
The SSIM is available in TF as the command 
\begin{mycodebox}
   tf.image.ssim(
      img1, img2, max_val, filter_size=11, filter_sigma=1.5, k1=0.01, k2=0.03
   )
\end{mycodebox} 
and can be readily called in a loss function to compare two images. 
See the command for all available parameters, such as Gaussian filter size and width to determine the region for the calculation of $\mu_i$ and $\sigma_i$.
TF also includes the command 
\begin{mycodebox}
   tf.image.ssim_multiscale(
      img1, img2, max_val, power_factors=_MSSSIM_WEIGHTS, filter_size=11,
      filter_sigma=1.5, k1=0.01, k2=0.03
   )
\end{mycodebox}
to implement the multiscale SSIM (MS-SSIM) developed by \cite{wang2003multiscale}.

Note that since the SSIM uses local averaging in the form of Gaussian filters, it can be considered a neighborhood measure, according to the classification developed by \cite{gilleland2009intercomparison,gilleland2010verifying}
and discussed in Section \ref{spatial_verificaton_sec}.

\subsection{Judging image similarity using pre-trained neural networks: perceptual loss}
\label{VGG_sec}

While SSIM and MS-SSIM are commonly used measures in the computer vision literature (see Section \ref{SSIM_sec}), they are not a perfect match for human perception.  The computer vision community has thus developed more sophisticated measures that more closely match human judgement of image similarity. 
A key idea is that rather than comparing two images pixel by pixel, it is often more meaningful to compare the presence and intensity of certain {\it patterns}, as represented in the feature maps of certain convolutional neural networks (CNNs), such as the well known VGG16 network \cite{simonyan2014very}.  Thus, one method to compare two images is to feed both images into a pretrained VGG16 network
and then compare their corresponding feature maps in the deeper convolutional layers of the VGG16 network (deeper layers represent more abstract, or high-level, patterns). These VGG16 feature map representations are often called {\it deep features} \cite{zhang2018unreasonable} or {\it deep feature maps} \cite{wang2020image}.
Ready-to-use code and detailed instructions on using VGG16 for image similarity are provided at 
\url{https://github.com/salmariazi/image_similarity_keras}.
The resulting image similarity measure can then be used as a loss function (called a {\it perceptual loss}) in a second neural network to be trained for an image translation task, such as style transfer or super-resolution \cite{johnson2016perceptual}.  

This is thus an example of a neural network with a loss function that calls another neural network during training.  While this may at first sound somewhat similar to the discriminator network in GANs, this set-up is quite different. Namely, the neural network used by a perceptual loss function is pretrained and its parameters are fixed, while the discriminator network used as loss in GANs is trained along with the generator network.  However, both set-ups share the fact that due to the very abstract loss function, it is hard to understand the training process of the neural network. Once trained though, the loss function is no longer used, and explainable AI methods \cite{ebert2020evaluation,McGovern2019-transparent} can be utilized to gain insights into either neural network type, just like for NNs that use traditional loss functions.


\section{Insights from the practical use of custom losses in environmental science}
\label{sec:recommendations}
\label{vignettes_sec}

This section provides insights and practical examples from practitioners in environmental science in the form of vignettes.\\

\begin{mdframed}
{\bf Vignette \#1: Insights from Christina Kumler}\\[0.2cm]
{\bf Application:} Detecting tropical and extratropical cyclones\\[0.2cm]
We developed four U-net models to identify either tropical cyclones, or both tropical and extratropical cyclones \cite{kumler2020tropical}.  Inputs to the U-nets consisted of two global images (covering the whole Earth): one from satellite measurements and one from the Global Forecasting System (GFS), which is a process-based atmospheric model.  At most time steps, many cyclones are present over the globe.  We experimented with different loss functions for this task, which falls under the category of semantic segmentation.  We tested both traditional loss functions and those that emphasize the minority classes (tropical and extratropical cyclones, as opposed to non-cyclones, the majority class) by assigning a higher weight to these pixels.  Specifically, we tested cross-entropy, the Dice coefficient, Tversky coefficient, and focal loss.

Table \ref{fig:TC_ROI_table} below shows the best combination of loss function and U-net depth for four different tasks, where each task is defined by the choice of one predictor dataset and one label dataset.
In general, the loss function yielded a negligible difference in performance (not shown).  However, U-nets trained with different loss functions had a substantial difference in the Tversky coefficient (not shown).  Also, as shown in the table, each of the four loss functions was selected exactly once.  Thus, no single loss function was clearly the best choice across the four tasks.
Furthermore, even for different U-nets trained on the same task, the best loss function often changed with other hyperparameters, such as model depth (not shown).  In general, we recommend that multiple loss functions be tested and that the scores not used in the loss function be tracked as metrics.

\begin{table}[H] 
    \noindent \centering 
    \includegraphics[width = 12cm, keepaspectratio]{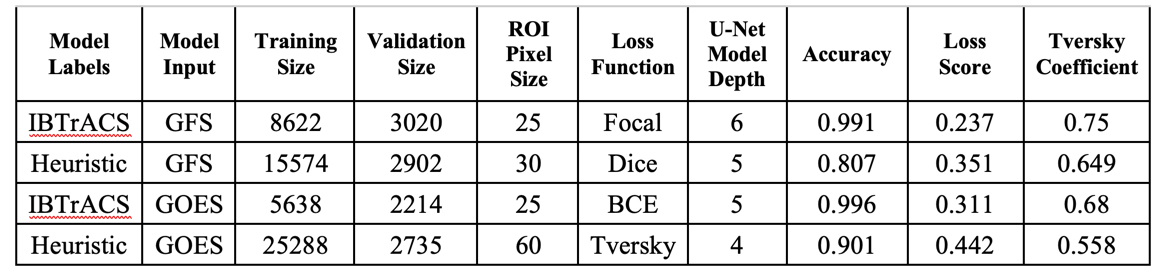} \\
    \caption{Details on the selected U-net for each task The Tversky coefficient was used, in addition to the accuracy and loss value, to evaluate how well the truth label regions and U-net identified regions overlapped one another. The pixel size indicates how large the cube around the cyclone was labeled in the input image. Adapted from Table 2 in \cite{kumler2020tropical}.}
    \label{fig:TC_ROI_table} 
\end{table}
\begin{figure}[H] 
    \noindent \centering 
    \includegraphics[width = 15cm, keepaspectratio]{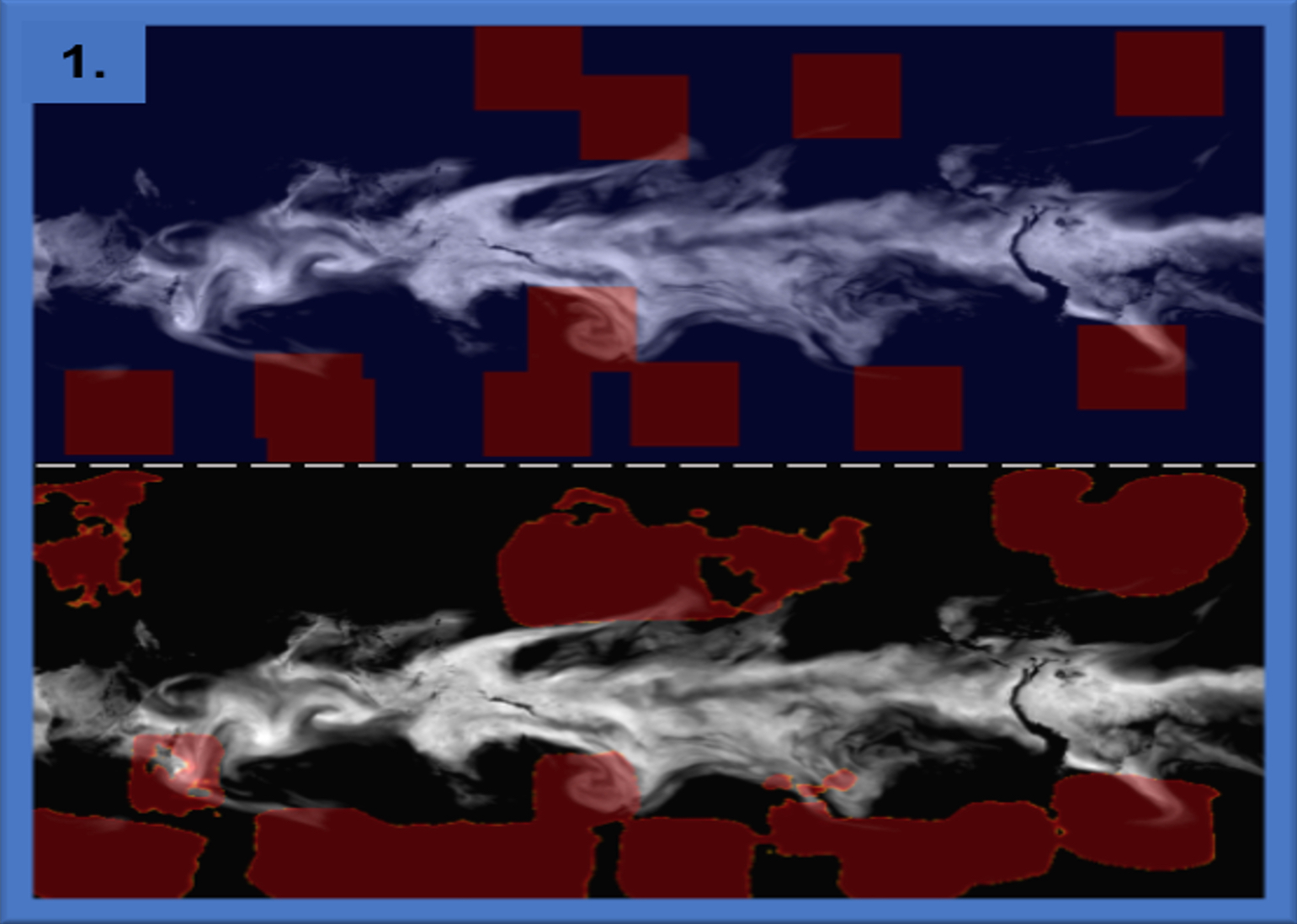} \\
    \caption{Sample results for cyclone detection.  The top panel shows total precipitable water (TPW) from the GFS (blue-white color scheme, with white corresponding to higher TPW), and the red boxes in the top panel show fixed-size bounding boxes around actual cyclones, from the label dataset.  The bottom panel also shows TPW from the GFS (same map), but the red boxes here show estimated cyclone locations, according to the U-net. Adapted from Figure 4 in \cite{kumler2020tropical}. \copyright American Meteorological Society. Used with permission.}
    \label{fig:TC_ROI_figure}  
\end{figure}
\end{mdframed}


\begin{mdframed}
{\bf Vignette \#2: Insights from Kyle Hilburn}\\[0.2cm]
{\bf Application:}  Generating synthetic radar imagery\\[0.2cm]
Figure \ref{fig:performance_diagram_K6} shows how weights were determined for the generalized exponential loss function given in Section \ref{save_and_load_model_subsec}, applied to the problem of generating synthetic radar reflectivity from GOES satellite observations using a performance diagram. When no weights are used (solid black line), Figure \ref{fig:performance_diagram_K6} shows that the model severely underpredicts high radar reflectivity values, which are rare in the dataset. Experiments were conducted, training the model with 45 different settings, where \emph{b} was varied from 1 to 5 by 0.5 and \emph{c} was varied from 1 to 5 by 1. The optimal weight settings were determined by minimizing the categorical bias \cite{wilks-statistical_2011} calculated at a set of reflectivity thresholds $i$. The best model was selected by taking $\underset{k}{\textrm{min}} \lbrace \underset{i}{\textrm{mean}} \lvert 1 -\textrm{ bias}_{i, k} \rvert \rbrace$,
where the subscript $i$ indexes the reflectivity threshold and $k$ indexes the model.  This approach identified weights $b$=5 and $c$=4 (solid red line), which provide 148x more weight for the highest values (60 dBZ). The weight has a sharper transition at high values than the weights implied by taking the inverse of the radar reflectivity PDF (solid blue line), which gives $b$=5 and $c$=1. More details on the GREMLIN model are provided in \cite{DevelopmentandInterpretationofaNeuralNetworkBasedSyntheticRadarReflectivityEstimatorUsingGOESRSatelliteObservations}.
\begin{figure}[H] 
    \noindent \centering 
    \includegraphics[width = \textwidth, keepaspectratio]{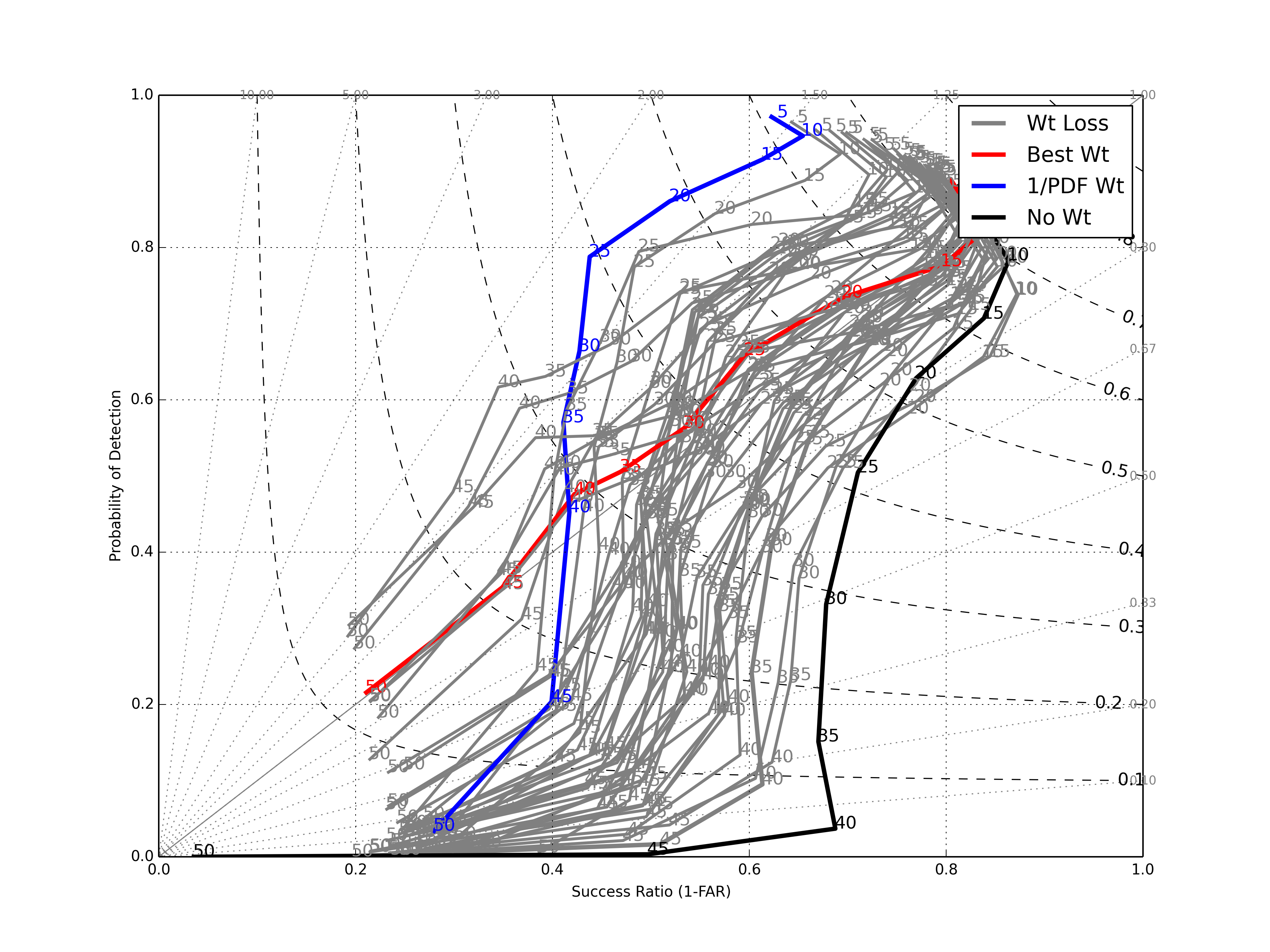} \\
    \caption{Performance diagram for composite radar reflectivity categories 5 to 50 dBZ in steps of 5 dBZ. The dashed black lines are critical success index and the gray dotted lines are categorical bias. The solid gray lines show the model performance for 45 different weight settings. Results with no weights are given by the solid black line, the optimally determined weights given by the solid red line, and the weights implied by the PDF by the solid blue line.  Adapted from Figure 5 in \cite{ebert2020evaluation}. \copyright American Meteorological Society. Used with permission.} \label{fig:performance_diagram_K6}
\end{figure}
\end{mdframed}

\begin{mdframed}
{\bf Vignette \#3: Insights from Ryan Lagerquist} \\ [0.2cm]
{\bf Application: }Physically consistent and skillful radiative flux \\ [0.2cm]
Figures \ref{fig:flux_results_unconstrained} and \ref{fig:flux_results_constrained} show the results of using a physically unconstrained and constrained loss function, respectively, for radiative transfer.  The exact loss functions are discussed in Section \ref{subsec:physical_constraints}.  Both figures show the attributes diagram for each flux component \textendash \, $F_\downarrow$, $F_\uparrow$, and $F_{\textrm{net}}$ \textendash \, based on independent testing data.  The attributes diagram \cite{Hsu1986} is a reliability curve with additional reference lines and histograms.  The reliability curve itself (the orange, purple, or green line) plots the prediction on the $x$-axis and conditional mean observation on the $y$-axis.  For example, if the reliability curve includes the point (100 W m\textsuperscript{-2}, 150 W m\textsuperscript{-2}), this means that in cases where the prediction is 100 W m\textsuperscript{-2}, the mean observation is 150 W m\textsuperscript{-2}.  The diagonal grey line is the one-to-one line, where prediction = conditional mean observation.  The horizontal and vertical grey lines are not explained here, for the sake of brevity, except to note that their position corresponds to the climatological value (\textit{i.e.}, mean flux value over the whole dataset).  The blue shading is the positive-skill area, where the model has a lower Brier score (MSE for binary classification) than a climatological model \textendash \, \textit{i.e.}, one that predicts climatology for every example.  Lastly, the histograms show the distribution of the predictions and observations.  In a perfect attributes diagram, the reliability curve follows the one-to-one line and the two histograms are identical.  Other than the left tail of the reliability curve for single-layer cloud (\textit{c.f.} Figures \ref{fig:flux_results_unconstrained}b and \ref{fig:flux_results_constrained}b), the physically constrained loss function improves predictions of all three flux components, especially $F_{\textrm{net}}$.

\begin{figure}[H]
	\noindent \centering
	\includegraphics[width = \textwidth, keepaspectratio]{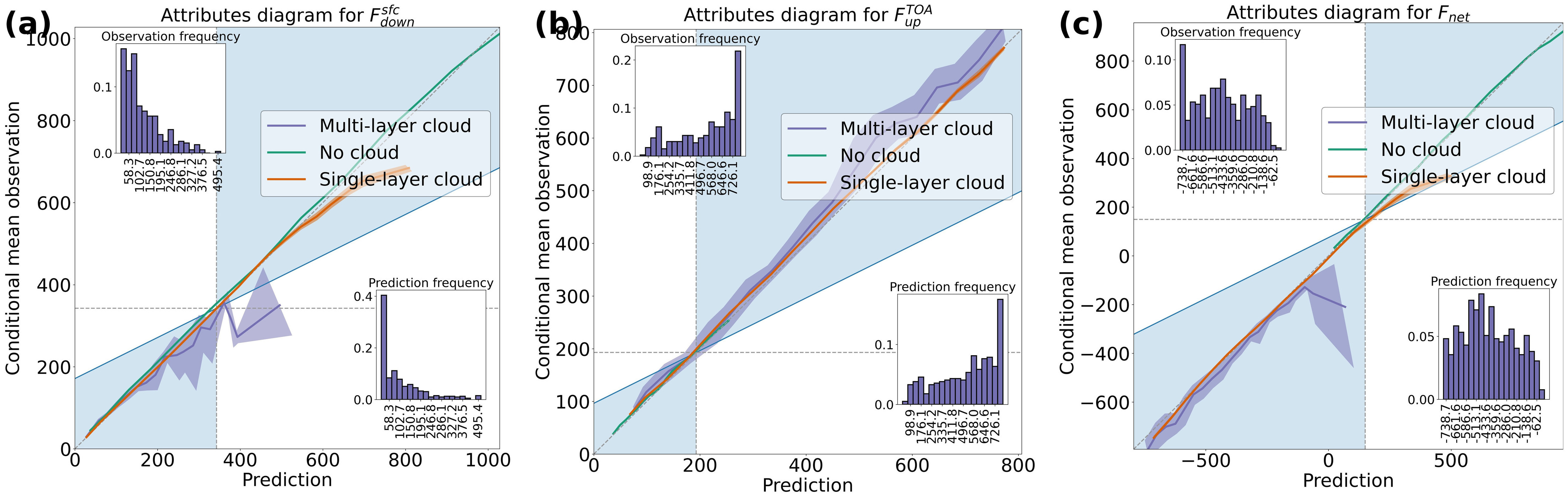} \\
	\caption{Attributes diagram for each flux component, using a loss function without physical constraints (Equation \ref{equation:net_flux_loss} without the $F_{\textrm{net}}$ term).  The testing data are split into three regimes: profiles with no cloud, those with one liquid-cloud layer, and those with multiple.  For each reliability curve, the solid line is the mean, while the shaded envelope is the 99\% confidence interval, determined by bootstrapping with 1000 replicates.  Adapted from Figure 7 in \cite{Lagerquist2021_radiation}.}
	\label{fig:flux_results_unconstrained}
\end{figure}

\begin{figure}[H]
	\noindent \centering
	\includegraphics[width = \textwidth, keepaspectratio]{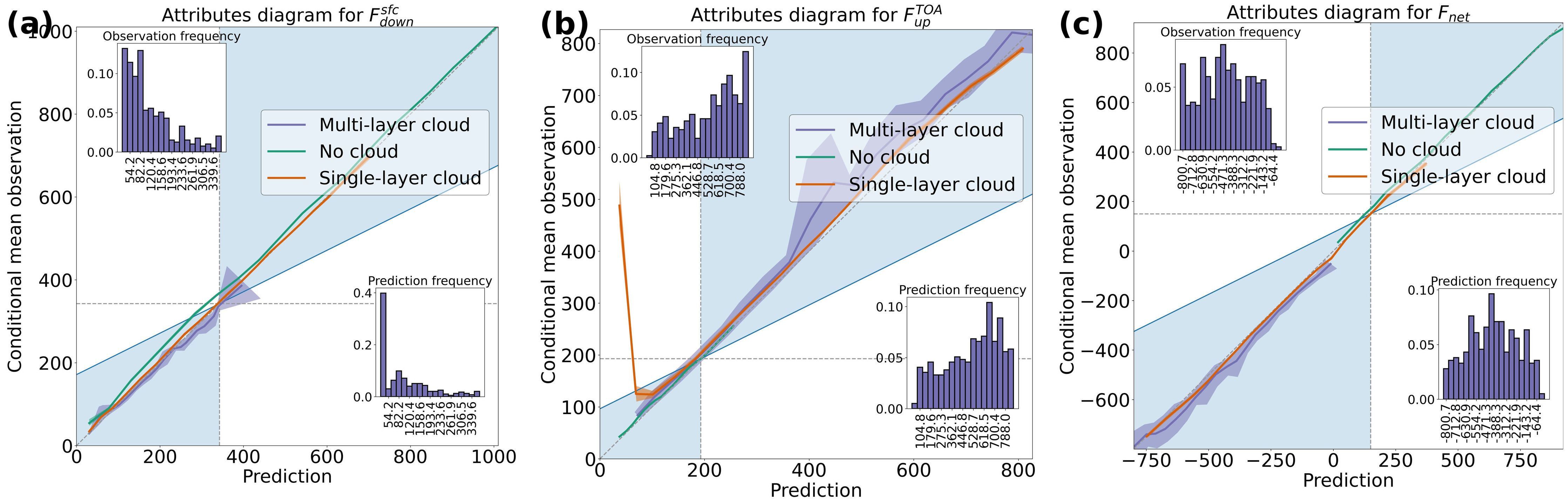} \\
	\caption{Same as Figure \ref{fig:flux_results_unconstrained} but for neural network with physically constrained loss function (Equation \ref{equation:net_flux_loss}).  Adapted from Supplemental Figure S26 in \cite{Lagerquist2021_radiation}.}
	\label{fig:flux_results_constrained}
\end{figure}
\end{mdframed}

\section{Concluding Comments}
\label{conclusions_sec}

We hope that this document provides a comprehensive resource for environmental scientists on the following topics:
\begin{itemize}
\item
   Basic introduction on how to implement custom loss functions, including common pitfalls; 
\item
   Ideas for many different function types that can be used in loss functions, such as NN layer functions (e.g., pooling and convolution), commonly available functions from image processing (e.g., Sobel operators, edge detectors), and those from signal processing (e.g., FFT);
\item
   Implementation of the fractions skill score as loss function;
\item
   Advanced loss function concepts to consider to address specific problems, such as loss functions to enforce physical constraints, adaptive loss, focal loss, robust loss, and perceptual loss.
\end{itemize}
In summary, we hope that this document enables readers to be creative and to implement a large range of new loss functions 
that teach neural networks what scientists in the environmental sciences truly care about, thus yielding NNs that better suit their applications.

\section{Acknowledgements}

We would like to thank many colleagues for stimulating discussions on related topics, including Chris Slocum (NOAA), Elizabeth Barnes (Atmospheric science at CSU), John Haynes (CIRA), Yoo-Jeong Noh (CIRA), Jason Apke (CIRA), and Kate Musgrave (CIRA). 
Ebert-Uphoff acknowledges support for this work by the National Science Foundation under NSF grant ICER-2019758 (NSF AI institute) and NSF grant OAR-1934668.  Hilburn acknowledges support for this work by NOAA's GOES-R Program through award NA19OAR4320073.
We would like to thank NOAA RDHPCS for access to the Fine Grain Architecture System on Hera, without which this research would not have been possible.


\newpage
\bibliographystyle{ieeetr}
\bibliography{references}


\end{document}